%% file: main.tex
\documentclass{article}

\usepackage[final]{neurips_2024}

\usepackage[utf8]{inputenc} %
\usepackage[T1]{fontenc}    %

\input{preamble}

\input{macros}

\usepackage{float}
\usepackage{mathtools}
\usepackage{bm}
\usepackage{amsmath}
\usepackage{amssymb}
\usepackage{graphicx}
\usepackage[inkscapelatex=false]{svg}
\usepackage{kantlipsum}
\usepackage{nicefrac}
\usepackage{dsfont}
\usepackage[labelfont=bf, format=plain]{caption}
\usepackage{wrapfig}

\title{Flexible task abstractions emerge in linear networks with fast and bounded units}

\author{%
  Kai Sandbrink\footnotemark[1] \\
  Exp. Psychology, Oxford\\
  Brain Mind Institute, EPFL
  \And
  Jan P. Bauer\footnotemark[1] \\
  ELSC, HebrewU \\
  Gatsby Unit, UCL \\
  \And 
  Alexandra M. Proca\footnotemark[1] \\
  Department of Computing \\
  Imperial College London \\
  \AND
  Andrew M. Saxe \\
  Gatsby Unit, UCL \\
  \And
  Christopher Summerfield \\
  Exp. Psychology, Oxford
  \And
  Ali Hummos\footnotemark[1]\:\:\footnotemark[2] \\
  Brain and Cognitive Sciences \\
  MIT \\
}

\begin{document}
 
\maketitle

\footnotetext[1]{Equal contribution, randomized order}
\footnotetext[2]{Corresponding author, \texttt{ahummos@MIT.edu}}

\begin{abstract}
Animals survive in dynamic environments changing at arbitrary timescales, but such data distribution shifts are a challenge to neural networks. To adapt to change, neural systems may change a large number of parameters, which is a slow process involving \textit{forgetting} past information. In contrast, animals leverage distribution changes to segment their stream of experience into tasks and associate them with internal task abstractions. Animals can then respond \textit{flexibly} by selecting the appropriate task abstraction. 
However, how such flexible task abstractions may arise in neural systems remains unknown.
Here, we analyze a linear gated network where the weights and gates are jointly optimized via gradient descent, but with neuron-like constraints on the gates including a faster timescale, nonnegativity, and bounded activity. We observe that the weights self-organize into modules specialized for tasks or sub-tasks encountered, while the gates layer forms unique representations that switch the appropriate weight modules (task abstractions). We analytically reduce the learning dynamics to an effective eigenspace, revealing a virtuous cycle: fast adapting gates drive weight specialization by protecting previous knowledge, while weight specialization in turn increases the update rate of the gating layer. Task switching in the gating layer accelerates as a function of curriculum block size and task training, mirroring key findings in cognitive neuroscience. We show that the discovered task abstractions support generalization through both task and subtask composition, and we extend our findings to a non-linear network switching between two tasks. 
Overall, our work offers a theory of cognitive flexibility in animals as arising from joint gradient descent on synaptic and neural gating in a neural network architecture.
\end{abstract}

\section{Introduction}
Humans and other animals show a remarkable capacity for flexible and adaptive behavior in the face of changes in the environment. Brains leverage change to discover latent factors underlying their sensory experience  \citep{gershman_learning_2010, yu_adaptive_2021, castanon_mixture_2021}: they segment the computations to be learned into discrete units or `tasks'.
After learning multiple tasks, low-dimensional task representations emerge that are abstract (represent the task invariant to the current input) \citep{bernardi_geometry_2020} and compositional \citep{tafazoli_building_2024}. The discovery of these useful task abstractions relies on the temporal experience of change, and in fact, brains struggle when trained on randomly shuffled interleaved data \citep{flesch_comparing_2018, beukers_blocked_2024}.

In contrast, while artificial neural networks have become important models of cognition, they perform well in environments with large, shuffled datasets but struggle with temporally correlated data and distribution shifts. To adapt to changing data distributions (or `tasks'), neural networks rely on updating their high-dimensional parameter space, even when revisiting previously learned tasks – leading to catastrophic forgetting \citep{mccloskey_catastrophic_1989, hadsell_embracing_2020}. One way to limit this forgetting is through task abstractions, either provided to the models \citep{hummos_gradient-based_2024} or discovered from data \citep{hummos_thalamus_2023}. In addition, adapting a model entirely by updating its weights is data-intensive due to high dimensionality. Task abstractions simplify this process by allowing updates to a low-dimensional set of parameters, which can be switched rapidly between known tasks, and recomposed for new ones. However, despite the advantages of task abstractions, simple algorithms for segmenting tasks from a stream of data in neural systems remain an open challenge.

This paper examines a novel setting where task abstractions emerge in a linear gated network model with several neural pathways, each gated by a corresponding gating variable. We jointly optimize the weight layer and gating layer through gradient descent, but impose faster timescale, nonnegativity, and bounded activity on the gating layer units, making them conceptually closer to biological neurons. 
We find two discrete learning regimes for such networks based on hyperparameters, a \textit{flexible} learning regime in which knowledge is preserved and task structure is integrated flexibly, and a \textit{forgetful} learning regime in which knowledge is overwritten in each successive task. In the flexible regime, the gating layer units align to represent tasks and subtasks encountered while the weights separate into modules that align with the computations required. Later on, gradient descent dynamics in the gating layer neurons can retrieve or combine existing representations to switch between previous tasks or solve new ones. Such flexible gating-based adaptation offers a parsimonious mechanism for continual learning and compositional generalization \citep{butz_learning_2019, hummos_thalamus_2023, qihong_lu_episodic_2024, schug_discovering_2024}.
Our key contributions thus are as follows:
\begin{itemize}
    \item We \textbf{describe flexible and forgetful modes of task-switching} in neural networks and \textbf{analytically identify the effective dynamics} that induce the flexible regime.
    \item The model, to our knowledge, is \textbf{the first simple neural network model that benefits from data distribution shifts and %
    longer task blocks rather than interleaved training
    } \citep{flesch_comparing_2018,beukers_blocked_2024}. We also provide a direct comparison to human behavior where task switching accelerates with further task practice \citep{steyvers_large-scale_2019}.
    \item We \textbf{generalize our findings to fully-connected deep linear networks}. We find that differential learning rates and regularization on the second layer weights are necessary and sufficient for earlier layers to form task-relevant modules and later layers to implement a gating-based solution that selects the relevant modules for each task. 
    \item We \textbf{extend our findings to non-linear networks}. As a limited proof of concept, we embed such a layer in a non-linear convolutional network learning two-digit classification tasks. 
\end{itemize}

\section{Related work}
Cognitive flexibility allows brains to adapt behavior in response to change \citep{miller_integrative_2001, egner_principles_2023, sandbrink_modelling_2024}. Neural network models of cognitive flexibility frequently represent knowledge for different tasks in distinct neural populations, or modules, which then need to be additionally gated or combined~\citep{musslick_rationalizing_2021, yang_task_2019}. Several models assumed access to ground truth task identifiers and used them to provide information about the current task demands to the network \citep{kirkpatrick_overcoming_2017, masse_alleviating_2018, yang_task_2019, wang_contextual_2022, driscoll_flexible_2024, hummos_gradient-based_2024}. Indeed having access to task identifiers facilitates learning, decreases forgetting, and enables compositional generalization \citep{yang_task_2019, masse_flexible_2022, hummos_gradient-based_2024}. Such works sidestep the problem of discovering these task representations from the data stream. 

Other models train modular neural structures end-to-end, such as mixture-of-experts \citep{jacobs_task_1991, jordan_hierarchical_1994, tsuda_modeling_2020}, or modular networks \citep{andreas_neural_2016, kirsch_modular_2018, goyal_recurrent_2019}.
A fundamental issue is the `identification problem' where different assignments of experts to tasks do not significantly influence how well the model can fit the data, making identification of useful sub-computations via specialized experts difficult \citep{geweke_interpretation_2007}. Practically, this results in a lack of modularity with tasks learned across many experts \citep{mittal_learning_2020} or expert collapse, where few experts are utilized \citep{krishnamurthy_improving_2023}. Recent work used a surprise signal to allow temporal experience to adapt learning \citep{barry_fast_2022}. Our model proposes simple dynamics that benefit from the temporal structure to assign sub-tasks to modules.

Our work builds on the theoretical study of \textit{linear} networks which exhibit complex learning dynamics, but are analytically tractable \citep{Saxe13Exactsolutionsnonlinear, saxe_mathematical_2019}. Prior work examined how gating alleviates interference \citep{saxe_neural_2022}, but gating was static and provided as data to the network. We generalize this line of work by showing how appropriate gating emerges dynamically.
More recently, \citet{shi_learning_2022} analyzed specialization of a linear network with multiple paths when tasks are presented without blocking and gates, and \citet{lee_why_2024} studied the effects of a pretraining period. We consider continual learning with a blocked curriculum. 
\citet{schug_discovering_2024} proved that learning a linear number of (connected) task module combinations is sufficient for compositional generalization to an exponential number of module combinations %
in a modular architecture similar to ours.
Instead, we explicitly study the interaction between task learning and gating variable update dynamics.

\begin{wrapfigure}[21]{r}{0.36\textwidth}
   \centering
   \includegraphics[width=0.36\textwidth]{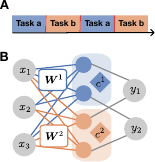} %
   \caption{\textbf{The open-ended learning setting and the modeling approach.} \textbf{A.} Example of the blocked curriculum with two tasks. \textbf{B.} Neural Task Abstraction (NTA) model updates $\bm{W}^p$ through gradient descent, but also the gating variables $c^p$, leading to task abstractions emerging in the gating layer.}
   \label{fig:approach}
\end{wrapfigure}
\section{Approach}

\subsection{Setting}

We formulate a dynamic learning problem consisting of %
$M$ distinct tasks. At each time step $t$ the network is presented with an input and output pair ($\bm{x}(t), \bm{y} ^{\star m} (\bm{x}(t))$) sampled from the current task $m$. Tasks are presented in blocks lasting a period of time $\tau_B$ before switching to another task sequentially (\cref{fig:approach}A). Models are never given the task identity $m$ or task boundaries. 

Specifically, we consider a multitask teacher-student setup in which each task is defined by a teacher $\bm{W}^{\star m}$,
which generates a ground truth label $\bm{y}^{\star m}=\bm{W}^{\star m}\bm{x}$ with a Gaussian i.i.d. input $\bm{x}$ at every point in time.
We randomly generate the teachers to produce orthogonal responses to the same input. While orthogonal tasks simplify theoretical analysis, we generalize to non-orthogonal tasks in \cref{app:nonortho_teachers}.

\subsection{Model}

We study the ability of linear gated neural networks \citep{Saxe13Exactsolutionsnonlinear, saxe_neural_2022} to adapt to teachers sequentially. We use a one-layer linear gated network with $P$ student weight matrices $\bm{W}^{p} \in \mathbb{R}^{d_{\text{out}} \times d_\text{in}}$, together with $P$ scalar variables $c^p \in \mathbb{R}$ which gate a cluster of neurons in the hidden layer (\cref{fig:approach}B).

The model output ${\bm y}\in \mathbb{R}^{d_\text{out}}$ reads
\begin{equation}
\label{eq:lcs}
\tag{1, NTA}%
    \bm{y} = \sum_{p=1}^P c^{p} \bm{W}^{p}\bm{x}. 
\end{equation}
Since the $c^p$ variables will learn to reflect which task is currently active, we refer to their activation patterns as \textit{task abstractions}. We refer to a student weight matrix together with its corresponding gating variable as a \textit{path}.

We refer to this architecture as the Neural Task Abstractions (NTA) model when the following two conditions are met during training: first, 
we update both the weights $\bm{W}^p$ and the gating variables $c^p$ via gradient descent, but on a regularized loss function $\mathcal{L} = \mathcal{L}_{\text{task}} +  \mathcal{L}_{\text{reg}}$. 
Second, we impose a shorter timescale for the gates $\tau_c$ than for the weights $\tau_w$, i.e. $\tau_c < \tau_w$ (although this condition becomes unnecessary if the task is sufficiently high-dimensional, see \cref{app:heavy_teachers}).

The task loss is a mean-squared error $\mathcal{L}_\text{task}=\nicefrac{1}{2}\sum_i^{d_{\text{out}}} %
\langle(y_i^{\star m} - y_i)^2\rangle$ where the average is taken over a batch of samples. 
The regularization loss contains two components $\mathcal{L}_{\text{reg}} =\lambda_{\text{norm}} \, \mathcal{L}_{\text{norm}} +  \lambda_{\text{nonneg}} \, \mathcal{L}_{\text{nonneg}}$ weighted by coefficients $\lambda_\text{norm}, \lambda_\text{nonneg}$. 
The normalization term bounds gate activity, $\mathcal{L}_{\text{norm}} = \nicefrac{1}{2} \, (||\bm{c}||_k - 1)^2$, where we consider $k=1, 2$. The nonnegativity term favors positive gates $\mathcal{L}_{\text{nonneg}} = \sum_{p=1}^P \max(0, -c^p)$. Together, these regularizers incentivize the model to function as an approximate mixture model by driving solutions towards any convex combination of students without favoring specialization and reflect constraints of biological neurons (see \cref{app:regularization} for details).

Assuming small learning rates (\textit{gradient flow}), this approach implies updates of 

\noindent
\begin{minipage}{.5\textwidth}
\begin{equation}\tau_c \nicefrac{d}{dt}\,{c^p}=- \,\nabla_{c^p}\,\mathcal{L}%
,\nonumber
\end{equation}
\end{minipage}%
\begin{minipage}{.5\textwidth}
\begin{equation}\tau_w\nicefrac{d}{dt}\,{\bm{W}^p}=- \,\nabla_{\bm{W}^p}\,\mathcal{L}%
\nonumber
\end{equation}
\end{minipage}

where $\tau_c$ and $\tau_w$ are time constants of the model parameters. We initialize $\bm{W}^p$ as i.i.d. Gaussian with small variance $\sigma^2/d_{\text{in}}$, $\sigma=0.01$ and $c^p=\nicefrac{1}{2}$.

Code for model and simulations at: \href{https://github.com/aproca/neural_task_abstraction}{https://github.com/aproca/neural\_task\_abstraction}

\begin{figure}[t]
\centering
\includeinkscape[width=\textwidth]{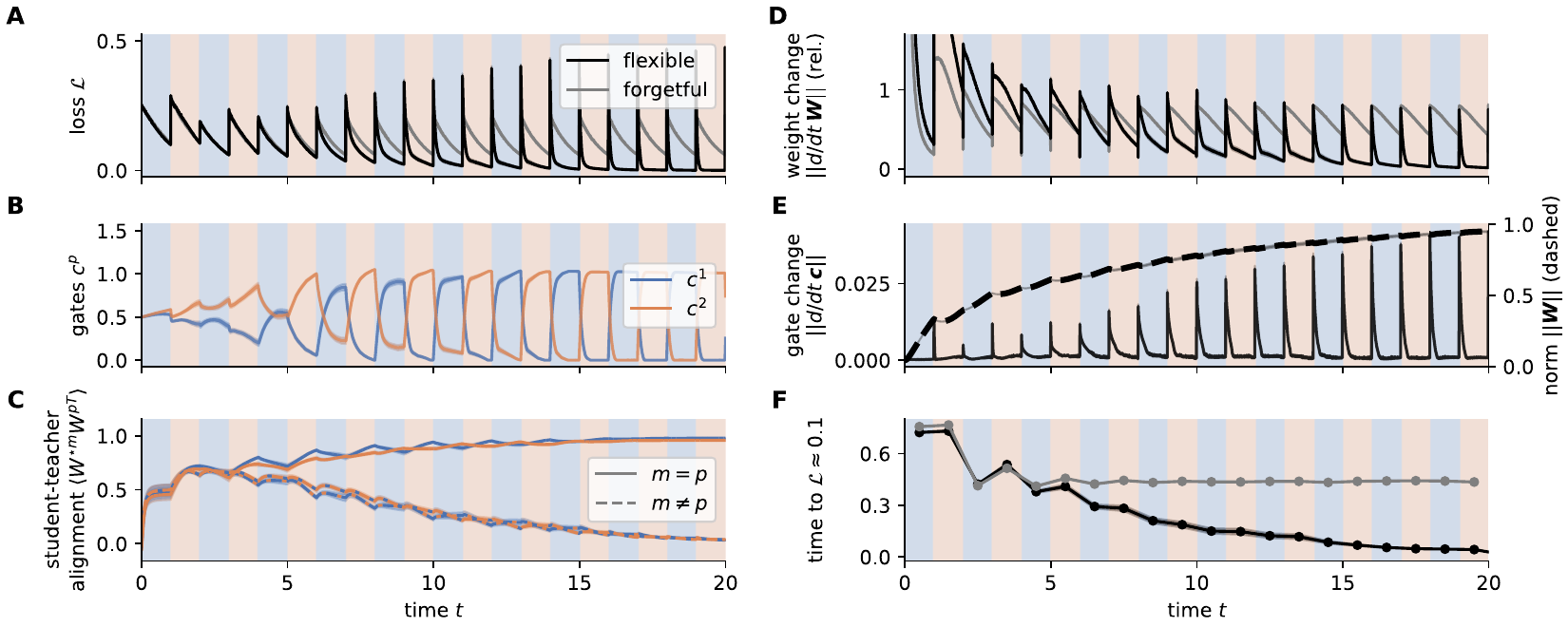}
\caption{\textbf{Joint gradient descent on gates and weights enables fast adaptation through gradual specialization.} Learning on the blocked curriculum from \cref{fig:approach} with $\tau_c=0.03$, $\tau_w=1.3$, and block length $\tau_B=1.0$. $x$-axis indicates time as multiples of $\tau_B$. (\textit{Black}) Flexible NTA model \cref{eq:lcs}, (\textit{gray}) forgetful NTA model with $\tau_c=\tau_w$ and $\lambda_{\text{nonneg}} = \lambda_{\text{norm}} = 0$. Simulation averaged over 10 random seeds with standard error indicated. \textbf{A.} Loss of both models over time. \textbf{B.} Gate activity of flexible NTA. \textbf{C.} Student-teacher weight alignment  $\bm{W}^{\star m} \bm{W}^{p \T}$, normalized and averaged over rows (cosine similarity) for each student-teacher pair. 
\textbf{D., E.} Norm of updates to $\bm{W}^p$ and $\bm{c}$. \textit{Dashed}: norm of students correlating with update size of $\bm{c}$. \textbf{F.} Time to $\mathcal{L}_\text{task} = 0.1$ for both models over blocks.}
\label{fig:main}
\end{figure}

\section{Task abstractions emerge through joint gradient descent}

\begin{wrapfigure}{r}{0.6\textwidth}
    \centering
        \includegraphics[width=0.6\textwidth]{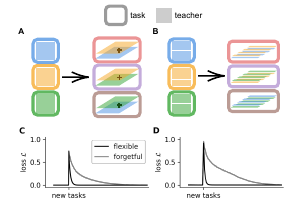}
        \caption{\textbf{Flexible model generalizes to compositional tasks.} \textbf{A.} Task composition consists of new tasks that sum sets of teachers previously encountered. \textbf{B.} Subtask composition consists of new tasks that concatenate alternating rows of sets of teachers previously encountered. Loss of models trained on generalization to task composition (\textbf{C.}) and subtask composition (\textbf{D.}) for the flexible (\emph{black}) and forgetful (\emph{gray}) NTA. `New tasks' indicates the start of the generalization phase when the task curriculum is changed to cycle through the compositional tasks.}
    \label{fig:generalization}
\end{wrapfigure}

We train the model with fast and bounded gates on $M=2$ alternating tasks (\cref{fig:approach}A) and use $P=2$ paths for simplicity (for the $P\lessgtr M$ case, see \cref{fig:generalization} and \cref{app:repr_cost}). As a baseline, we compare to the same model but without gate regularization and timescales difference.

Both models reach low loss in early blocks, but only flexible NTA starts to adapt to task switches increasingly fast after several block changes (\cref{fig:main}A,F). Analyzing the model components reveals what underlies this accelerated adaptation (\cref{fig:main}C,D): in early blocks of training, zero loss is reached through re-aligning both students $\bm{W}^p$ to the active teacher $\bm{W}^{\star m}$ in every block, while the gates $c^p$ are mostly drifting (\cref{fig:main}B). Reaching low loss is furthermore only achieved towards the end of a block. Later, the weights stabilize to each align with one of the teachers (\cref{fig:main}C,D), and the appropriate student is selected via gate changes (\cref{fig:main}B), reducing loss quickly. The rate at which gates change correlates with the alignment and magnitude $||\bm{W}^p||$ of the learned weights (\cref{fig:main}C,E). Overall, this points towards a transition between two learning regimes: first, learning happens by aligning student weight matrices with the current teacher, which we call \textit{forgetful} because it overwrites previous weight changes. Later, as the weights specialize, the learned representations $\bm{W}^p$ can be rapidly selected by the gates according to the task at hand, reflecting adaptation that is \textit{flexible}. Only the model equipped with fast and bounded gates (flexible NTA) is able to enter this flexible regime (\cref{fig:main}A,F). 

Next, we verify that the task abstractions in the gating variables are general, in the sense that they support compositional generalization. We consider two settings that begin by training a model with three paths on three teachers A, B, and C in alternating blocks, and then training on novel conditions. In \textit{task composition}, the novel conditions are the teachers' additive compositions A+B, A+C, B+C (\cref{fig:generalization}A), we see that the flexible NTA model trains on these combinations much faster (\cref{fig:generalization}C). In \textit{subtask composition}, the novel conditions are combinations of the rows of different teachers, i.e. we break the teachers A, B, C into rows and select from these rows to compose new tasks.
(\cref{fig:generalization}B). 
In the subtask composition case, we use a more expressive form of the gates in the model that can control each row of the student matrices $\mathbf{W}^p$ individually. %
We find that, in the flexible regime, the model quickly adapts to the introduction of compositional tasks, while the forgetful model with regularization removed
does not (\cref{fig:generalization}C,D). For more details and extended analysis, see \cref{app:generalization}. 

We devote the next section to identifying what factors support the flexible regime of learning.

\section{Mechanisms of learning flexible task abstractions}
\label{sec:mechanism}

We observed in \cref{fig:main} that simultaneous gradient descent on weights and gates converges to a flexible regime capable of rapid adaptation to task changes. But what mechanisms facilitate this solution? We here leverage the linearity of the model to identify the effective dynamics in the SVD space of the teachers, in which we describe the emergence and functioning of the flexible regime.

\subsection{Reduction to effective 2D model dynamics} %

\begin{figure}[t]
    \centering
    \includegraphics[width=0.85\textwidth]{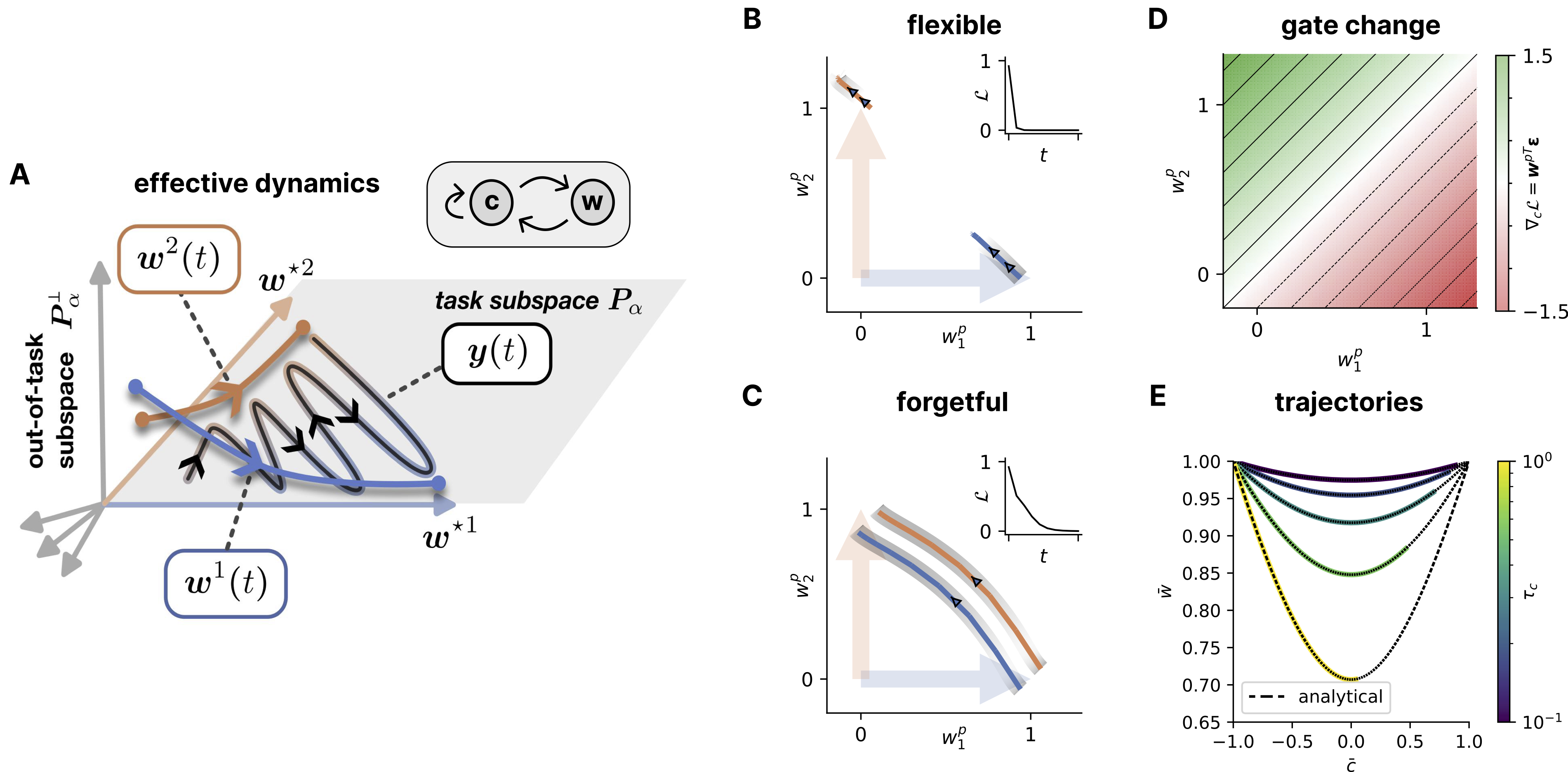}
    \caption{\textbf{Mechanism of gradual task specialization in effective 2D subspace.} \textbf{A.} Sketch of the reduced model and dynamic feedback. Out-of-subspace students gradually align to teacher axes. %
    \textbf{B.} Trajectories of student weight matrices (\textit{blue, orange}) in the teacher subspace during complete adaptation following a context switch from teacher 1 to teacher 2 in the flexible regime. \textit{Gray stripes} indicate associated gate activation. The student weight matrices move little. 
    \textbf{C.} Like (\textbf{B}), but for the forgetful regime. Student weight matrices entirely remap and gates do not turn off. 
    \textbf{D.} Gradient of the task loss on $c^p$ as a function of the weight alignment. \textbf{E.} Trajectories in the specialization subspace as a function of gate timescale for values $\tau_c=0.1, 0.18, 0.32, 0.56, 1.00$ comparing (\textit{color}) simulations and (\textit{dashed black}) analytical predictions from exact solutions under symmetry in the flexible regime. Simulations begin from initial conditions of complete specialization and separation $w^p_m=\delta_{pm}$, $c^p = \delta_{p1}$ and follow a complete adaptation from teacher 1 to teacher 2 over the course of a block, reaching $\mathcal{L}_\text{task}<10^{-2}$ for all $\tau_c$. }
    \label{fig:mechanism}
\end{figure}

For simplicity, we consider the case with only $M=P=2$ teachers and students. We take a similar approach to \cite{Saxe13Exactsolutionsnonlinear}, and project the student weights into the singular value space of the teachers for each mode $\alpha$ individually, yielding a scalar $w^{p}_{m\alpha} \coloneqq \bm{u}^{\star m \T}_\alpha \bm{W}^p \bm{v}^{\star m}_\alpha$.  Each pair of components $\alpha$ thus reduces to a 2D state vector $\bm{y}=c^{1}\bm{w}^{1}+c^{2}\bm{w}^{2}\in\mathbb{R}^2$, where we stack $w_{m}^p$ along the index $m$ and omit $\alpha$  in the following for readability (\cref{fig:mechanism}A).  %
A similar projection is possible in terms of the row vectors of both teachers (\cref{app:vector_ws}). 

The essential learning dynamics of the system can therefore be described as

\noindent
\begin{minipage}{.5\textwidth}
\begin{equation}%
\tau_w \, \dt\bm{w}^{p}=c^{p}\left(\bm{y}^{\star m} -\bm{y}\right),\label{eq:dw_toy}
\end{equation}
\end{minipage}%
\begin{minipage}{.5\textwidth}
\begin{equation}%
\tau_c \, \dt c^{p}=\bm{w}^{p\T}\left(\bm{y}^{\star m}- \bm{y}\right) - \lambda \, \nabla _{c^p} \mathcal{L}_{\text{reg}}
.\label{eq:dc_toy}
\end{equation}
\end{minipage}

\noindent where $\bm{y}^{\star m}$ describes the output of the currently-active teacher $m$. In \cref{app:reduction}, we show analytically and through simulations that this reduction is exact when gradients are calculated over many samples.

\subsection{Specialization emerges due to self-reinforcing feedback loops}

The flexible regime is characterized by students that are each attuned to a single teacher (\cref{fig:mechanism}B), whereas in the forgetful regime, both students track the active teacher together (\cref{fig:mechanism}C). We can describe this difference by studying the specialization of the students. We define this by considering the difference in how represented the teachers are in the two paths: for teacher 1, $\bar{w}_1 \coloneqq w_{m=1}^{p=1} - w_{m=1}^{p=2}$ and, for teacher 2, $\bar w_2 \coloneqq w_{m=2}^{p=2} - w_{m=2}^{p=1}$. %
Similarly, a hallmark of the flexible regime are separated gates. Together, this defines the specialization subspace

\begin{minipage}{.5\textwidth}
\begin{equation}%
\bar{w} \coloneqq \frac{1}{2} (\bar w_1 + \bar w_2),
\end{equation}
\end{minipage}%
\begin{minipage}{.5\textwidth}
\begin{equation}%
\bar{c} \coloneqq c^1 - c^2
\label{eq:spec}
\end{equation}
\end{minipage}
The system is in the flexible regime when absolute values of $\bar w$ and $\bar c$ are high (approaching 1), and in the forgetful regime when they are low (around 0).
In this section, we study the emergence of the flexible regime through self-reinforcing feedback loops, with specialized students and normalizing regularization leading to more separated gates, and separated gates in turn leading to more specialized students. In each subsection, we first describe the effect of the feedback loops on the paths individually, before considering the combined effect on specialization. Without loss of generality, we consider cases where the student $p$ specializes to teacher $m=p$.

\subsubsection{Specialized students and regularization encourage fast and separated gates}

We first investigate the influence of $\bm{w}^{p}$ on $\dt{c^{p}}$.
From the gate update in \cref{eq:dc_toy}, we get 
\begin{equation}
\tau_{c}\dt c^{p}=\varepsilon_{1}\,\bm{w}^{p\T}\bm{w}^{\star1}\:+\:\varepsilon_{2}\,\bm{w}^{p\T}\bm{w}^{\star2} \quad - \quad \nabla_{c^p} \mathcal{L}_{\text{reg}} ,\label{eq:magnitude_alignment}
\end{equation}
where we decomposed the error $\bm{\varepsilon}\coloneqq\bm{y}^{\star m} - \bm{y}$ into the teacher basis as coefficients
$\varepsilon_{m}\coloneqq\bm{\varepsilon}^{\T}\bm{w}^{\star m}$. The feedback between students and gates enters here in two terms,
as can be seen by expressing $\bm{w}^{p\T}\bm{w}^{\star m}=||\bm{w}^{p}||\,||\bm{w}^{\star m}||\,\cos\left(\angle\left(\bm{w}^{p},\bm{w}^{\star m}\right)\right)$,
where $\angle$ denotes the angle between two vectors. As observed in \cref{fig:main}, both the \textit{alignment} between students and teachers $\cos\left(\angle\left(\bm{w}^{p},\bm{w}^{\star m}\right)\right)$ and the \emph{magnitude} of the students $||\bm{w}^{p}||$ control the gate switching speed.

As the vectors $\bm{w}^p$ are formed from the students' singular values, they scale proportionally to the bare matrix entries $W^p_{ij}$ for random initialization (Marcenko-Pastur distribution). Early in learning, the small initialization will therefore attenuate gate changes by prolonging their effective timescale $\tau_{c} / ||\bm{w}^{p}||$ (or equivalently, lower their learning rate). 

As we demonstrate in \cref{fig:mechanism}D, these effects apply
to both activation and inactivation of the gates, depending on the direction of the current error $\bm{\varepsilon}^{\T}\bm{w}^{\star m}\gtrless 0$. 

The regularization in the system introduces a feedback loop between $c^{1}$ and $c^2$. In practice, the system quickly reaches a regime where both gates $c^p$ are positive. In this case, the regularization term using the $L^1$-norm becomes $\nabla_{c^p} \mathcal{L}_{\text{reg}} \propto \sum_{p'} c^{p'}  - 1$, reaching a minimum along the line $\sum_{p'} c^{p'} = 1$. In order to minimize the regularization loss, the upscaling of one gate $c^p$ past $0.5$ will result in the downscaling of the other gate $c^{p'}$, and vice versa.
We note that this regularization term does not favor specialization by itself since the network can also attain zero loss in the unspecified forgetful solution with, for instance, $c^1 = c^2 = 0.5$. 

The above dynamics mean that differences in student alignment separate the gates, as described by%

\begin{equation}
\tau_c \dv{\bar c}{t} = \bar w_1  \, \varepsilon_1 - \bar w_2 \, \varepsilon_2
\label{eqn:ldbarc}
\end{equation}

We therefore see that the differences in gate activation are driven by the difference in specialization in the two components $\bar w_1$ and $\bar w_2$ and corresponding error components $\varepsilon_1$ and $\varepsilon_2$. Since the error components are of opposite sign following a context switch, $\dv{\bar c}{t}$ is maximized when the students are maximally specialized.

\subsubsection{Flexible gates protect learned specialization}

We now study the influence of $c^p$ on $\dt{\bm{w}^p}$. The gates allow for a switching mechanism that does not
require a change in the weights. When continuing gradient descent
on all parameters, however, \cref{eq:dw_toy} will also entail a finite update
to the wrong student.

If we Taylor-expand to second order, this update reads

\begin{equation}
\label{eq:taylor_dw_ddt}
\tau_w \dt \bw^{p}  \: \simeq \: c^{p}\bm{\varepsilon} \: + \: \frac{1}{2}\left(\Bigl(\dt c^{p}\Bigr)\,\be+c^{p}\,\Bigl(\dt\be\Bigr)\right)\, dt.
\end{equation}

The first summand of the second term reflects the protection that arises from changes in gating $\dt c^{p}=\bw^{p \T}\be$: a task switch to $\bm{y}^{\star m}=(0,1)^\T$ incurs an error $\bm{\varepsilon}\propto(-1,1)^{\T}$. For a specialized, but now incorrect student $\bm{w}^{p}\propto\left(1,\,0\right)^{\T}$, this term becomes $\dt c^{p}=\bw^{p \T}\be <0$
for the incorrect student. Together with the decreasing error in the last term $\dv{t}\be$, this reduces the student update from the leading-order first term $c^p \be$. Importantly, this protection effect grows over training as the student's contribution to the error $\bw^{p \T}\be$ increases.

Alongside protection, flexible gates also accelerate specialization, as can be seen by considering $\bm{w}$ in specialization space,  

\begin{equation}
\tau_w \dv{\bar{w}}{t} = \frac{1}{2}  \bar c  \, (\varepsilon_1 - \varepsilon_2)
\label{eqn:ldbarw}
\end{equation}

This equation shows that the students specialize through two factors: the difference in error between the two components $\varepsilon_1-\varepsilon_2$, and the difference in gate activation $\bar c = c^1-c^2$.

\subsection{Exact solutions to the learning dynamics describe protection and adaptation under symmetry in the flexible regime}

In this section, we study exact solutions to the learning dynamics in \cref{eqn:ldbarc} and \cref{eqn:ldbarw} to describe the behavior of the model as it switches between tasks when it is already in the flexible regime. To solve the differential equations, we require the condition of symmetry where %
$\bar w = \bar w_1 = \bar w_2$. %
This condition is approximately true for specialized states in the flexible regime (see \cref{fig:assumptions}), and its persistence follows as long as $\varepsilon_1 = - \varepsilon_2 $ holds or in the limit of strong $L^1$ regularization.%

We use the method presented in~\cite{shi_learning_2022} to solve the resulting dynamics of the ratio between the expressions for $\dv{\bar{c}}{t}$ and $\dv{\bar{w}}{t}$

\begin{gather}
\frac{\tau_c}{\tau_w} \dv{\bar{c}}{\bar{w}} = 2\,\frac{ \bar{w} \, ( \varepsilon_1 - \varepsilon_2)}{\bar{c} \, (\varepsilon_1 - \varepsilon_2)}  %
\label{eqn:shietaldiffeq}
\end{gather}

\noindent which is a separable differential equation that can be solved up to an integration constant  (see \cref{sec:solvingexactsolssupp}).  Plugging in initial conditions that correspond to complete specialization in the flexible regime $\bar{c}(0)=\bar{w}(0)=1$, we obtain the exact dynamics of $\bar{w}$ as a function of $\bar{c}$ over the course of a block,

\begin{equation}
    \bar{w} = \sqrt{1-\frac{1}{2}\frac{\tau_{c}}{\tau_w} \left(1-\bar{c}^2\right)}.
\end{equation}

This analytical solution accurately describes adaptation in the flexible regime (\cref{fig:mechanism}E).  The relationship highlights the role of a shorter gate timescale $\tau_c$ in protecting the student's knowledge. Learning that comes from both students specializing towards the current teacher occurs outside of this specialization space and becomes more important for low $\tau_c$ (see \cref{sec:cocomponent}).

\section{Quantifying the range of the flexible regime across block length, regularization strength, and gate speed}

\begin{wrapfigure}{r}{0.6\textwidth}
    \centering
\includeinkscape[width=.6\textwidth]{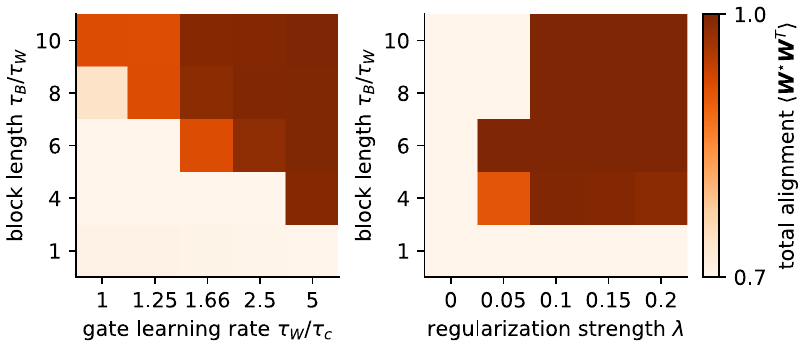}
        \caption{\textbf{Model specialization emerges as a function of block length, gate learning rate, and regularization strength.} The colorbar indicates total alignment (cosine similarity) between all sets of students and teachers considered collectively.}
    \label{fig:hypersearch}
\end{wrapfigure}

To assess the roles of block length, regularization, and fast gate timescale (inverse gate learning rate) %
in establishing the flexible regime, we run two grid searches over the gate learning rate/regularization strength and block length each task is trained on, keeping the total time trained constant (such that models trained on shorter block lengths are trained over more block switches but equal amounts of data). For each set of hyperparameters we compute the total alignment (cosine similarity) between the entire concatenated set of teachers and students as a single overall measure of specialization in the network weights at the end of learning. We identify the boundaries of the flexible regime where specialization emerges in our model, dependent on block length, gate timescale, and regularization strength (\cref{fig:hypersearch}). A priori, the block length dependence is surprising, as one might expect additional time spent in a block to be reversed by the equally-long subsequent block. However, we show in \cref{app:block-length} that gating breaks this time-reversal symmetry, and specialization grows with block length $\tau_{B}$ for fixed overall learning time $t$.

\section{Inducing the flexible regime in a deep fully-connected neural network}

Our NTA model uses a low-dimensional gating layer that gates computations from student networks. We sought to understand the necessity and role of this structure by considering a more general form of the model in a deep linear network with no architectural assumptions. Based on the analysis and results so far (\cref{fig:mechanism}D,\ref{fig:hypersearch}), we impose regularization and faster learning rates on the second layer of a 2-layer fully-connected network. 
Behaviorally, this network also shows the signatures of the flexible regime with adaptation accelerating with each task switch experienced (\cref{fig:gating_deep_mono_metrics}A).

\begin{wrapfigure}{r}{0.41\textwidth}
\centering
\includeinkscape[width=.41\textwidth]{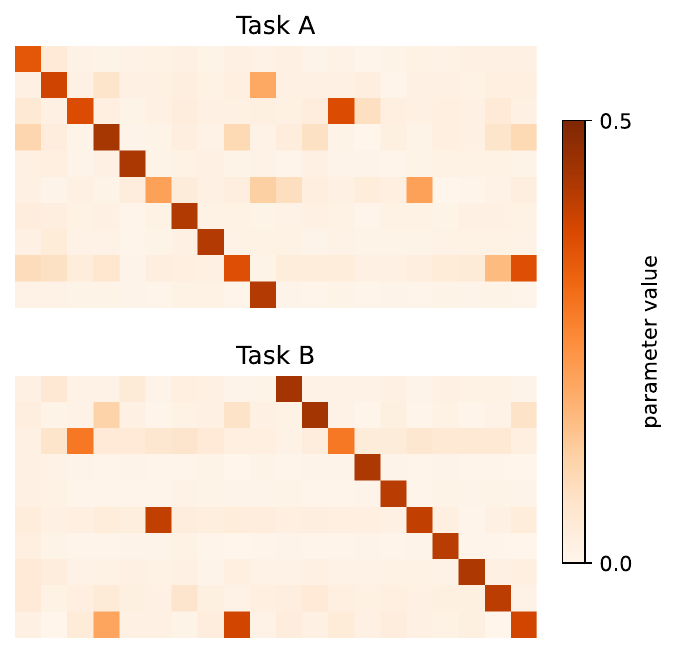}
\caption{\textbf{Task-specialized gating emerges in the second layer of a 2-layer network with faster second-layer learning rate and regularization.} The sorted second layer weights at the last timestep of two different task blocks (one seed). }
\label{fig:gating_deep_mono_sorted_gates}
\end{wrapfigure}

To quantify specialization and gating behavior, we compute the cosine similarity between each row of the first hidden layer and the teachers and use this to sort the network into two students that align to the teachers they match best. We also permute the second layer columns to match the sorting of the first layer.
We then visualize the specialization of the sorted first hidden layer using the same measures as in the original NTA model.
We also take the mean of each sorted student's second hidden layer to be its corresponding gate.
Using this analysis, we find emergent gating in the second layer (\cref{fig:gating_deep_mono_metrics}B) and specialization in the first (\cref{fig:gating_deep_mono_metrics}C). Adaptation to later task switches takes place primarily in the second layer (\cref{fig:gating_deep_mono_metrics}E). 

By visualizing the sorted second hidden layer of the fully-connected network at the last timestep of two different task blocks, we indeed observe distinct gating behavior along the diagonal, specialized for each task (\cref{fig:gating_deep_mono_sorted_gates} for one seed). We compare this to the same fully-connected network trained without regularization which remains in the forgetful learning regime. We include visualizations of the unsorted second hidden layer for fully-connected networks arriving at both the gating and non-gating solutions  (\cref{fig:gating_deep_mono_unsorted} for ten seeds), as well as the sorted second hidden layer (\cref{fig:gating_deep_mono_tenseeds} for ten seeds) as supplement. \cref{app:theory_fullyconnected_gating} discusses the potential for multiplicative gates to emerge in fully-connected architectures.

\section{Flexible remapping of representations in nonlinear networks in two MNIST tasks}

\begin{figure}[b]
    \centering
    \includegraphics[width=\textwidth]{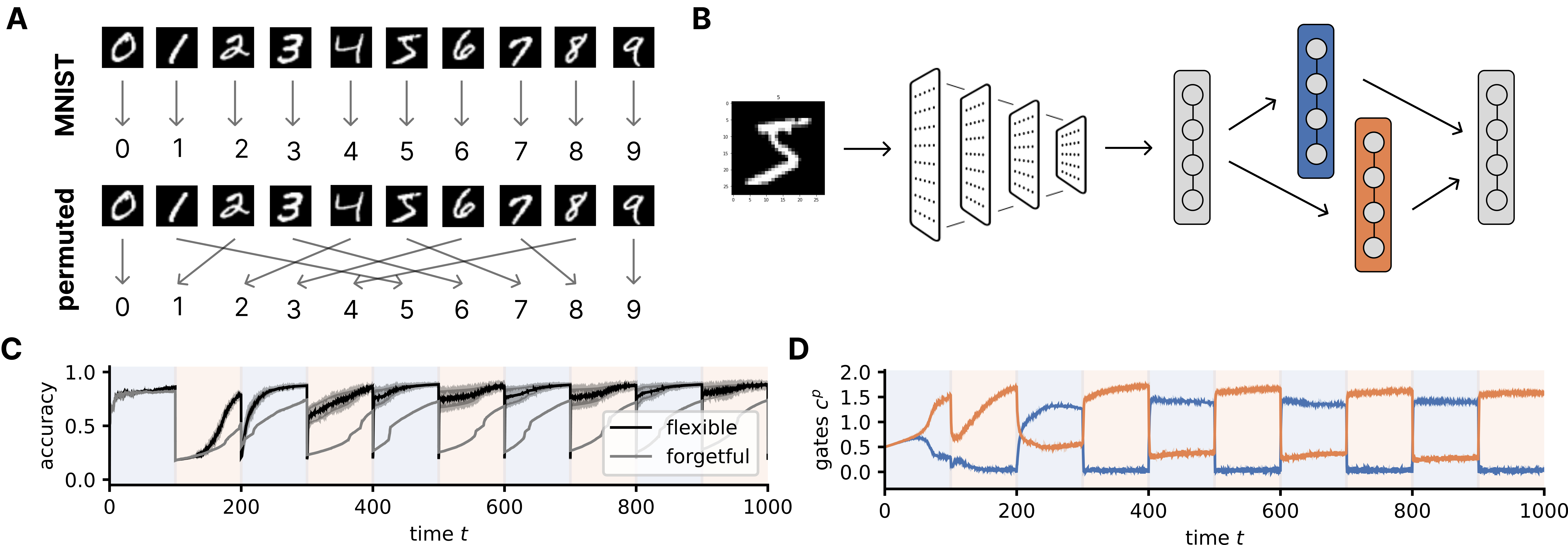}
    \caption{\textbf{Learning flexible neural task abstractions in a nonlinear character recognition setting. A.} We formulate two tasks, the original and a permuted version of MNIST. \textbf{B.} We embed the NTA system into a larger pretrained convolutional neural network architecture. \textbf{C.} Accuracy reached on the MNIST test set as a function of time for both (\textit{black}) the NTA network and (\textit{gray}) the original CNN. The two tasks are presented sequentially in blocks for both (\textit{blue shading}) MNIST and (\textit{orange shading}) the permuted version. \textbf{D.} The activation of the two gating units as a function of time. We show mean and standard error with 10 seeds. %
    }
    \label{fig:mnist}
\end{figure}

We next study whether NTA also works in larger, nonlinear systems. As a proof of concept, we investigate whether NTA can help a neural network switch between two nonlinearly-transformed versions of the MNIST dataset \citep{deng2012mnist}. %
The first task is the conventional MNIST task. The second is a permuted version of MNIST where the image of a digit is sorted based on its parity according to the function $y \to \lfloor y / 2 \rfloor + 5 \times (y \% 2)$, where $\%$ is the modulo operation (see \cref{fig:mnist}A). We pre-train a convolutional neural network (CNN) on MNIST to learn useful representations, achieving about 90\% accuracy on the test set.  %
We then train an NTA system beginning from the final hidden layer representations that feeds into the same sigmoid nonlinearity (see \cref{fig:mnist}B). %
We again induce the flexible regime using regularization and fast timescales, and contrast performance with a forgetful model
(see \cref{app:mnist}). We find that the flexible model learns to recover its original accuracy quickly after the first task switches whereas the forgetful one needs to continuously re-learn the task, as evaluated on the MNIST test set (\cref{fig:mnist}C). The activity in the gating units reflects selective activity (\cref{fig:mnist}D). %
To further test the range of NTA, we examine how much these results depend on the orthogonality of the task space by formulating two tasks based on real-world groupings of clothing in fashionMNIST \citep{xiao2017/online} that have different amounts of shared structure. We find that rapid task switching occurs in both settings at a similar speed (\cref{sfig:fashion_mnist}).

\section{Relations to multi-task learning in humans}

Our model captures several aspects of human learning. %
Humans update task knowledge in proportion to how well each probable task explains current experience \citep{castanon_mixture_2021}. Analogously in our model, weight updates are gated with the corresponding gating variable, whose activity in turn reflects %
how well the weights behind it capture the target computation.

\begin{wrapfigure}{r}{0.68\textwidth}
    \centering
\includeinkscape[width=.68\textwidth]{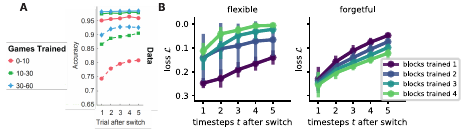}
        \caption{\textbf{Comparing performance after a task switch in humans and NTA model.} \textbf{A.} \cite{steyvers_large-scale_2019} report performance of humans learning two alternating tasks (CC BY-NC-ND 4.0 license). \textbf{B.} After a block switch, loss comparison between the flexible (\textit{left}) and the forgetful (\textit{right}) NTA model shows opposite trends with further training on switching speed. Bars are standard error with 10 seeds.}
    \label{fig:task_switching_speed}
\end{wrapfigure}

Humans show faster task switching with more practice on the tasks involved. In our model, we saw that the gates change faster as weights specialize to the tasks, which facilitated faster adaptation after block switches.
NTA shows a qualitative fit (\cref{fig:task_switching_speed}) to humans trained on alternating tasks \citep{steyvers_large-scale_2019}. In contrast, a forgetful model shows a deceleration, possibly due to being far from optimal initialization \citep{dohare_loss_2024} after task switches.%

\section{Conclusions and future work}
This study demonstrates how task abstraction and cognitive flexibility can emerge in neural networks trained through joint gradient descent on weights and gating variables. Simple constraints on the gating variables induce gradient descent dynamics that lead to the \textit{flexible regime}. In this regime, the weights self-organize into task-specialized modules, with gating variables facilitating rapid task switching. Analytical reductions revealed a virtuous cycle: specialized weights enable faster gating, while flexible gating protects and accelerates weight specialization. This contrasts with the \textit{forgetful regime} in which knowledge is continually forgotten and re-learned after task switches. The constraints necessary for reaching the flexible regime are appropriate regularization, differential learning rates, and sufficient task block length. These mirror properties of biological neurons and beneficial learning settings identified in cognitive science.

The mechanistic understanding of how task abstraction arises in neural systems might bridge artificial neural networks and biological cognition, offering a foundation for future exploration of adaptive and compositional generalization in dynamic environments. While this study focuses on simple two-layer networks, the framework is applicable to other non-linear architectures such as recurrent networks or Transformer architectures. We see future work providing additional architectures and real-world applications of the framework.

\newpage
\section*{Author contributions}
All authors contributed to manuscript writing and conceptualization of the work.
AP conceptualized and implemented the experiments, and contributed to the idea, the theory, and the model simulator.
JB conceptualized and developed the theory, developed the model simulator, and contributed to the idea and experiments. 
KS conceptualized and developed the theory, conceived of the idea and the link to cognitive flexibility, and contributed to experiments.
AH directed the project, contributed to the idea and the experiments, and provided feedback on the theory.

\section*{Acknowledgements}
We thank Stefano Sarao Mannelli and Pedro A.M. Mediano for thoughtful discussions. We would also like to acknowledge and thank the organizers of the Analytical Connectionism Summer School, at which AP, JB, and KS first met. %
AH is funded by Collaborative Research in Computational Neuroscience award (R01-MH132172). AP is funded by the Imperial College London President's PhD Scholarship. KS is funded by a Cusanuswerk Doctoral Fellowship. JB is supported by the Gatsby Charitable Foundation (GAT3850). This work was supported by a Sir Henry Dale Fellowship from the Wellcome Trust and Royal Society (216386/Z/19/Z) to AS, and the Sainsbury Wellcome Centre Core Grant from Wellcome (219627/Z/19/Z) and the Gatsby Charitable Foundation (GAT3755).

\bibliography{references.bib}

\clearpage
\appendix

\input{appendix}

\clearpage

\end{document}

%% file: preamble.tex
\usepackage{amsmath}
\usepackage{amssymb}
\usepackage{amsthm}

\usepackage[
  colorlinks = true,
  citecolor  = black,
  linkcolor  = black,
  urlcolor   = black,
  unicode,
]{hyperref}

\usepackage{hyperref}
\usepackage[capitalise]{cleveref}

\usepackage{bm}
\usepackage{units}

\usepackage{babel}
\usepackage{physics}
\usepackage{comment}
\usepackage{url}            %
\usepackage{booktabs}       %
\usepackage{amsfonts}       %
\usepackage{nicefrac}       %
\usepackage{microtype}      %
\usepackage{xcolor}         %

\usepackage{chngcntr}
\usepackage{makecell}

%% file: macros.tex
\global\long\def\dt{\dv{}{t}}%
\global\long\def\ddt{\dv[2]{}{t}}%
\global\long\def\bw{\bm{w}}%
\global\long\def\be{\bm{\varepsilon}}%
\global\long\def\NTK{\mathsf{NTK}}%

\makeatletter
\DeclareRobustCommand{\T}{%
  {\mathpalette\@transpose{}}%
}
\newcommand*{\@transpose}[2]{%
  \raisebox{\depth}{$\m@th#1\intercal$}%
}
\makeatother

%% file: appendix.tex
\section*{Appendix}
\counterwithin{figure}{section}  %

\subsection*{Overview}
\label{app:appendix-overview}

We structure the Appendix as follows: 

We first provide additional material on our results in \cref{app:additional-detail}. 
\cref{app:reduction} derives the reduction to the 2D equivalent model. 
\cref{app:heavy_teachers} shows that even without a differential timescale between gates and weights, high-rank students will learn more slowly compared to gates. 
\cref{app:repr_cost} shows that networks with more paths than teachers will split their representations across paths unless a cost is associated with representation. \cref{app:deep-monolithic} provides simulations and derivations on how gating behavior emerges in an architecture without explicit pathways but with two layers, where one layer emergently takes on the role of gates, and the second layer becomes compartmentalized.
\cref{app:deep-monolithic} contains additional results for the fully-connected network.
\cref{app:acceleration-NTK} shows how the specialized representation incentivized by the virtuous cycle discussed in the main text leads to a faster reduction in loss compared to an unspecialized solution. 
\cref{app:block-length} provides an approximate theoretical explanation for the beneficial effect of long blocks towards specialization through symmetry breaking in an effective potential. 
\cref{app:exact-solutions} provides approximate closed-form solutions when operating in the flexible regime. 
\cref{app:generalization} provides detail on how the model generalizes to new tasks by leveraging existing abstractions. 
\cref{app:nonortho_teachers} shows that the flexible regime largely persists and slowly decays when the orthogonality assumption between teachers is relaxed. 
\cref{app:rebuttal_samples} shows that the model can adapt in a few-shot fashion after a block switch, extending the results from the main text where gradients are calculated on many samples. 

We then provide additional technical details in \cref{app:technical-detail}. 
In \cref{app:notation}, we provide a notation table. \cref{app:hyperparam_list} lists parameters used for simulations. \cref{app:regularization} discusses the regularization that we use, in particular why it does not incentivize a flexible over a forgetful solution. 
\cref{app:metrics} details on how we calculate alignments between teachers and students, both for the per-student and per-neuron gating models.
\cref{app:hypersearch_details} discusses how we choose model parameters.

\section{Additional details on main text}\label{app:additional-detail}
\subsection{Derivation of reduction to 2D equivalent model}\label{app:reduction}

We will here show that the dynamics of the model can be reduced to
an effective model that acts in a 2D space spanned by the singular teacher vectors across both tasks $m$.

Recall the task loss as the mean-squared error
\begin{align*}
    \mathcal{L}_{\text{task}} & = \frac{1}{2Bd_\text{out}} \sum_{b}^{B} \| \bm{y}^{\star m}_b - \bm{y}_b \|^2 \\ 
    & = \frac{1}{2Bd_\text{out}} \sum_{b}^{B} \| \bm{W}^{\star m} \bm{x}_b - \sum_p c^p \bm{W}^p \bm{x}_b \|^2
\end{align*}
for a batch $\bm{X}=\left( \bm{x}_{b}\right) _{b=1\ldots B}$ of size $B$.

Following the approach in \cite{Saxe13Exactsolutionsnonlinear}, we assume the input data is whitened, such that the batch average $\frac{1}{B}\bm{X}\bm{X}^{\T}\approx \bm{I}_{d_{\text{in}}}$, and the learning rate $\tau^{-1}$ is small (i.e., the \textit{gradient flow} regime). Then, the batch gradient
reads as a differential equation that simplifies as
\begin{align}
\label{eq:W-grad}
\tau_w \dt\bm{W}^{p} & =\frac{1}{Bd_{\text{out}}}c^{p}\left(\bm{W}^{\star m}\bm{X}-\textstyle \sum_{p'} c^{p'} \bm{W}^{p'} \bm{X}\right)\bm{X}^{\T}\\
 & \approx c^{p}\left(\bm{W}^{\star m}-\textstyle \sum_{p'} c^{p'} \bm{W}^{p'}\right).
\end{align}

For each teacher $m$ and singular value decomposition along a mode $\alpha$ %
($\bm{u}_{\alpha}^{\star m},s_{\alpha}^{\star m},\bm{v}_{\alpha}^{\star m }$),
we can project this equation to get

\begin{align}
\label{eq:apdx-reduction}
\tau_w \dt\Bigl(\underbrace{\bm{u}_{\alpha}^{\star m\T}\bm{W}^{p}\bm{v}_{\alpha}^{\star m}}_{s_{m,\alpha}^{p}}\Bigr) & =c^{p}\left(\bm{u}_{\alpha}^{\star m\T}\bm{W}^{\star m}\bm{v}_{\alpha}^{\star m}-\bm{u}_{\alpha}^{\star m\T}\,{\textstyle \sum_{p'}}c^{p'}\bm{W}^{p'}\bm{v}_{\alpha}^{\star m}\right)\\
 & = c^{p}\left(\phantom{U_{\alpha}^{\star m\T}}s^{\star m}_{\alpha}\phantom{V_{\alpha}^{\star m}}-\phantom{U_{\alpha}^{\star m\T}}{\textstyle \sum_{p'}}c^{p'}s_{m,\alpha}^{p'}\phantom{V_{\alpha}^{\star m}}\right).\nonumber 
\end{align}

The student singular vectors $\bm{u}_{\alpha}^{p}, \bm{v}_{\alpha}^{p}$ have been shown to align to those of the current teacher  $\bm{u}_\alpha^{\star m}, \bm{v}_{\alpha}^{\star m}$ early in learning \citep{atanasov_neural_2021}. After training on both teachers, the student can therefore be fully described in terms of the coefficients $\left\{s_{m,\alpha}^{p}\right\}_{m\alpha}$ in the basis spanned by the $\alpha$-singular vectors of both teachers, decoupling from the other singular value dimensions. 
If all singular vectors across two teachers $m$ are pairwise orthogonal, these projections form an orthogonal basis.
The components outside of this projection will have finite error in
all context and therefore exponentially decay to 0 \citep{braun_exact_2022}.

This reduction allows us to study learning in a simpler and more interpretable
model. %
For the case where $M=P=2$ which we consider here for simplicity, we can therefore
reinterpret each model $\alpha$-component as vectors $\bm{w}^{1}\equiv(s_{m=1,\alpha}^{p=1},\,s_{m=2,\alpha}^{p=1})^{\T}$
, $\bm{w}^{2}\equiv(s_{m=1,\alpha}^{p=2},\,s_{m=2,\alpha}^{p=2})^{\T}$
in $\mathbb{R}^{2}$:
\begin{align}
\bm{{y}} & =c^{1}\bm{w}^{1}+c^{2}\bm{w}^{2}\label{eq:reduced_model}\\
 & =c^{1}\left(\begin{array}{c}
s_{1,\alpha}^{1}\\
s_{2,\alpha}^{1}
\end{array}\right)+c^{2}\left(\begin{array}{c}
s_{1,\alpha}^{2}\\
s_{2,\alpha}^{2}
\end{array}\right)\nonumber 
\end{align}

and redefine the context-dependent target vector $\bm{y}^{\star m}$
accordingly. %

The reduced model follows the update equations 
\begin{align}
\tau_w \dt\bm{w}^{p} & =c^{p}\left(\bm{y}^{\star m}-{\bm{y}}\right)\label{eq:dw_toy_alt}\\
\tau_c \dt c^{p} & =\bm{w}^{p\T}\left(\bm{y}^{\star m}-{\bm{y}}\right).\label{eq:dc_toy_alt}
\end{align}

Notably, both updates depend on the full error term $\be\coloneqq\left(\bm{y}^{\star m}-{\bm{y}}\right)$
with both paths entering into ${\bm{y}}$. The first equation
moves the student in the direction of the current total misestimation
of the active teacher $\be$. The second equation changes the gating of
the current path according to the alignment of the path $\bm{w}^{p}$
to the current vectorial error, reflecting the contribution of the
path to the mismatch. 

\begin{figure}[H]
    \centering
\includeinkscape[width=\textwidth]{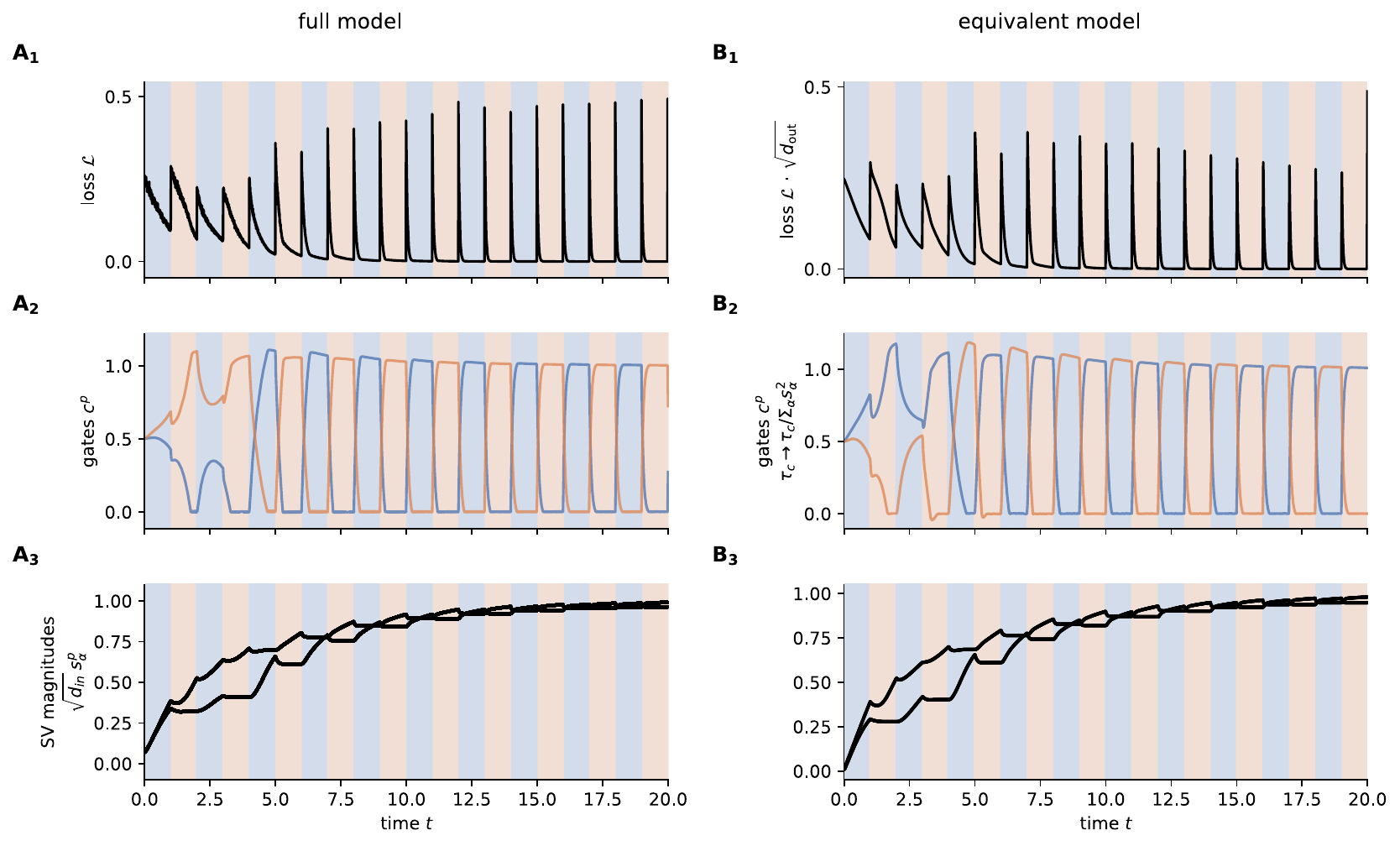}
    \caption{\textbf{Simulation of dynamics of full and reduced model.} The equivalent reduced model effectively captures the dynamics of the full model in terms of loss (\textbf{A$_1$.}, \textbf{B$_1$.}), gates (\textbf{A$_2$.}, \textbf{B$_2$.}), and singular value magnitude (\textbf{A$_3$.}, \textbf{B$_3$.}).}
    \label{fig:full-vs-toy}
\end{figure}

In \cref{fig:full-vs-toy}, we simulate the models side by side and show that the reduced model matches the dynamics of the full model.

\subsubsection{Reduction in terms of teacher row vectors}\label{app:vector_ws}

In the main text and the previous section, we consider a reduction that follows from projecting onto the eigenspace of the matrices. However, a similar reduction is possible by considering each row $\beta$ independently, and considering the row vectors of the two teachers $\left(\bm{w}_\beta^{\star m}\right)_m$ as a basis for that row. Like in the projection in terms of the eigenspace, the out-of-projection component of the students decays exponentially. We can then consider a single row of the teacher-student system to function as a mode $\alpha$ above, with a row of the student path $p$ becoming $\bm{w}^p = w^p_1 \, \bm{w}^{\star 1} + w^p_2 \, \bm{w}^{\star 2}$ so that we can write

\begin{align}
\bm{{y}} & =c^{1}\bm{w}^{1}+c^{2}\bm{w}^{2}\label{eq:vector_reduced_model}\\
 & =c^{1}\left(\begin{array}{c}
w_{1,\beta}^{1}\\
w_{2,\beta}^{1}
\end{array}\right)+c^{2}\left(\begin{array}{c}
w_{1,\beta}^{2}\\
w_{2,\beta}^{2}
\end{array}\right)\nonumber 
\end{align}

\noindent where we aggregate over rows $\beta$. This formulation only requires pairwise orthogonality between rows $\bm{w}_i^{\star 1} \cdot \bm{w}_i^{\star 2} = 0$ to fully decouple the dynamics of the system, but does not extend as elegantly to considering deeper students or low-rank solutions.

\subsection{High-dimensional students learn slower}\label{app:heavy_teachers}

\begin{figure}[htbp]
    \centering
\includeinkscape[width=0.4\textwidth]{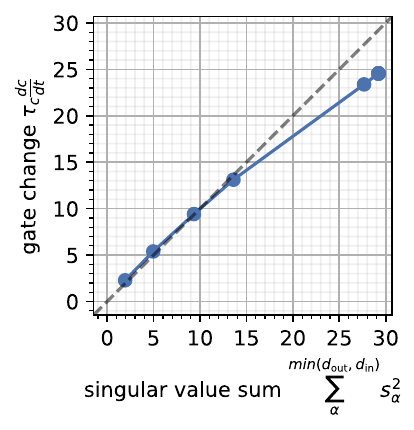}
    \caption{\textbf{High-dimensional students learn slower.} Gate change $\tau_c \dt c$ as a function of teacher dimensionality/rank (i.e., non-zero singular values). Weight scaling is chosen such that input and output components take unit scale, $y_i=\mathcal{O}(1)$,  $x_j=\mathcal{O}(1)$.}
    \label{fig:dimensionality}
\end{figure}

An intuition one might have for the model dynamics is that the weight
matrices comprise of more parameters and therefore may respond more slowly
under gradient descent. Here, we discuss the formal conditions under
which this indeed is the case. 

For simplicity, we consider a one-path model $\bm{y}=c \bm{W}\bm{x}$, with $\bm{y}\in\mathbb{R}^{d_{\text{out}}}$,
$\bm{x}\in\mathbb{R}^{d_{\text{in}}}$. We now choose the scaling $y_{i}=\mathcal{O}(1)$,
$x_{j}=\mathcal{O}(1)$ on input and output, which means
that entries do not depend on the respective vector dimensionalities.
This is a natural assumption, for example, if $y_{i}$ are label indicators
and $x_{j}$ are pixel brightness values of an image. Recall the model loss $\mathcal{L}_{\text{task}}=\nicefrac{1}{2}(\bm{y}^{\star m}-\bm{y})^2$.

Then, the batch-averaged $c$-gradient reads

\begin{align*}
\tau_c \dv{t}c=-\nabla_{c}\mathcal{L}_{\text{task}} & =\text{Tr}\left[(\bm{y}^{\star m}-\bm{W}\bm{x})\bm{x}^{\T}\bm{W}^{\T}\right],\\
 & =\text{Tr}\left[\bm{W}^{\star}\bm{x}\bm{x}^{\T}\bm{W}^{\T}-\bm{W}\bm{x}\bm{x}^{\T}\bm{W}^{\T}\right]\\
\langle\,\circ\,\rangle_B & \rightarrow\text{Tr}\left[\bm{W}^{\star}\bm{W}^{\T}-\bm{W}\bm{W}^{\T}\right]\\
 & \approx\text{Tr}\left[\bm{U}\bm{S}^{\star}\bm{V}\bm{V}^{\T}\bm{S}\bm{U}^{\T}-\bm{U}\bm{S}\bm{V}\bm{V}^{\T}\bm{S}\bm{U}^{\T}\right]\\
 & =\text{Tr}\left[\bm{S}^{\star}\bm{S}-\bm{S}^{2}\right]\\
 & =\sum_{\alpha}^{\text{min}(d_{\text{out}},\,d_{\text{in}})}\left(s_{\alpha}^{\star}s_{\alpha}-s_{\alpha}^{2}\right).
\end{align*}

Here, we used the Gaussian i.i.d. initialization of $\bm{x}$ to take
an expectation for large batch size ($\langle\bm{x}\bm{x}^{\T}\rangle_{B}=\bm{I}_{d_{\text{in}}}$),
the SVD of $\bm{W}=\bm{U}\bm{S}\bm{V}^{\T}$, orthonormalization of
singular vectors $\textbf{U}^{\T}\textbf{U}$, $\textbf{V}^{\T}\textbf{V}$,
and the cyclic property of the trace $\text{Tr}$. We also assumed in the fourth row that the student singular vectors have already undergone Silent Alignment
\citep{atanasov_neural_2021} to match the teachers, as discussed
in the main text. 

From the last row, we observe that the updates to $c$ tend to scale with the number of nonzero singular values, i.e. the rank of the teachers.

For the fan-in scaling ${W}_{ij}\sim\mathcal{N}(0,\,\sigma^{2}/d_{\text{in}})$
that is compatible with $x_{i}=\mathcal{O}(1)$, $y_{i}=\mathcal{O}(1)$,
we have $s_{\alpha}=\mathcal{O}(\sigma)$ independent of dimensionality (Marcenko-Pastur distribution).
If teacher and students are initialized according to this scaling,
students will respond relatively slower compared to gates as their
dimension $\text{min}(d_{\text{in}},d_{\text{out}})$ grows, as the
student gradient \cref{eq:W-grad} or the reduced form \cref{eq:apdx-reduction} does not
involve a sum that scales with dimensionality. 

\subsection{Representational cost in under-specified model}\label{app:repr_cost}
\begin{figure}[htbp]
    \centering
\includeinkscape[width=\textwidth]{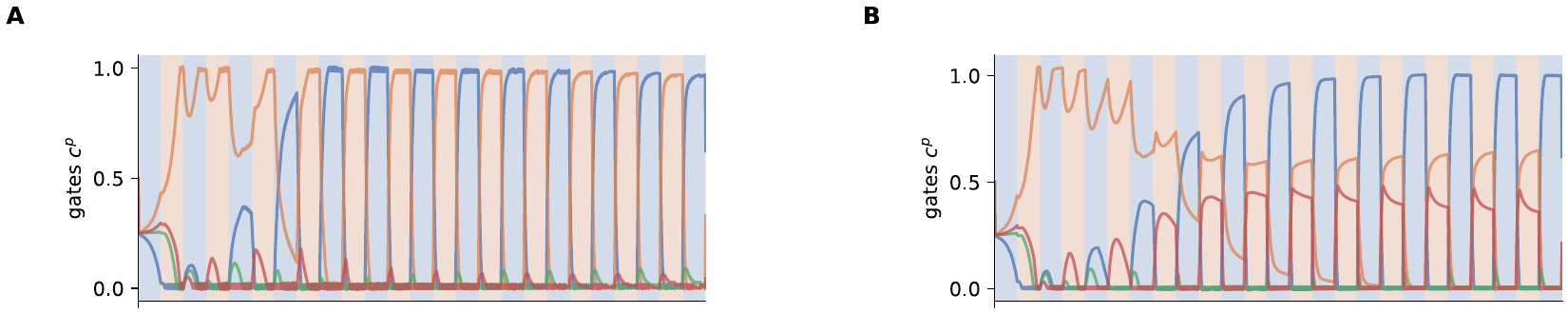}
    \caption{\textbf{Redundant paths become inactive when representation is costly.} Gating variables like in \cref{fig:approach}B, but with more paths than teacher tasks ($P=4 \,>\, M=2$). \textbf{A.} Only under representational cost on the weights, students that are preferably aligned due to the random initialization specialize to the $M=2$ teachers, whereas other gates decay to 0. \textbf{B.} Without representational cost, the model uses multiple paths for tasks and thus has multiple gates active at the same time for a single teacher.}
    \label{fig:repr_cost}
\end{figure}

In our initial \cref{eq:lcs}, we have introduced a model in which the number of paths $P$ of the architecture matches the number of available tasks. What happens if this match is not present? If the under-specified case $P<M$, the model's expressiveness hinders adaptation. It is however not clear what will happen in the over-specified case $P>M$. In absence of any regularization on the weights $\bm{W}^p$, the model will not devote only $P'=M<P$ paths to match the task. Rather, in accordance with the theory by \citet{shi_learning_2022}, the model will in general split its paths over the available tasks. This behavior is due to the absence of a ``representational cost'' of having multiple paths active at the same time. We find that this effect is reduced only when introducing an $L^2$-regularization $\frac{\lambda_W}{2Pd_{\text{in}}}\sum_{ijp}(W^p_{ij})^2$, $\lambda_W=0.77$. This term additionally penalizes weight magnitude leads to the decay of inactive paths. We show this behavior in \cref{fig:repr_cost}.

\subsection{Gating-based solution emerges in a fully-connected network}\label{app:deep-monolithic}

\begin{figure}[H]
    \centering
\includeinkscape[width=0.7\textwidth]{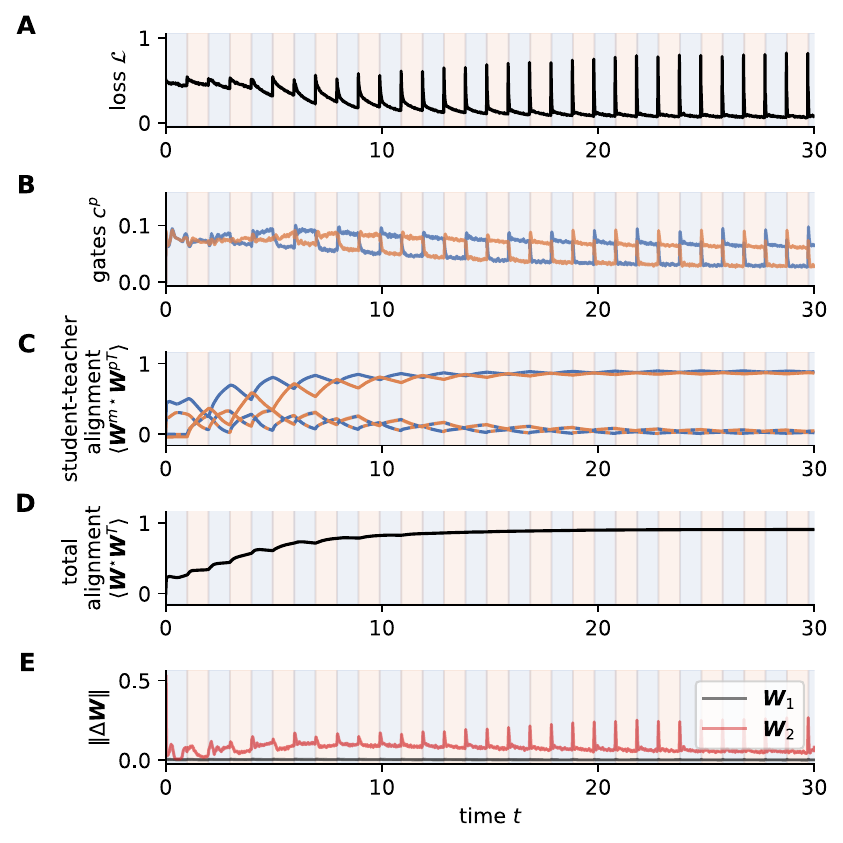}
    \caption{\textbf{Gating-based solution emerges in a fully-connected network with regularized second layer weights and a faster second layer learning rate.} \textbf{A.} Loss during learning. \textbf{B.} The dynamics of the sorted gating variables. \textbf{C.} Alignment between the sorted students in the first layer and the teachers. \textbf{D.} Total alignment between the entire set of teachers and students. \textbf{E.} The norm of the gradient of the first (\emph{black}) and second (\emph{red}) hidden layer of the fully-connected network.}
    \label{fig:gating_deep_mono_metrics}
\end{figure}

\begin{figure}[H]
    \centering
\includeinkscape[width=0.4\textwidth]{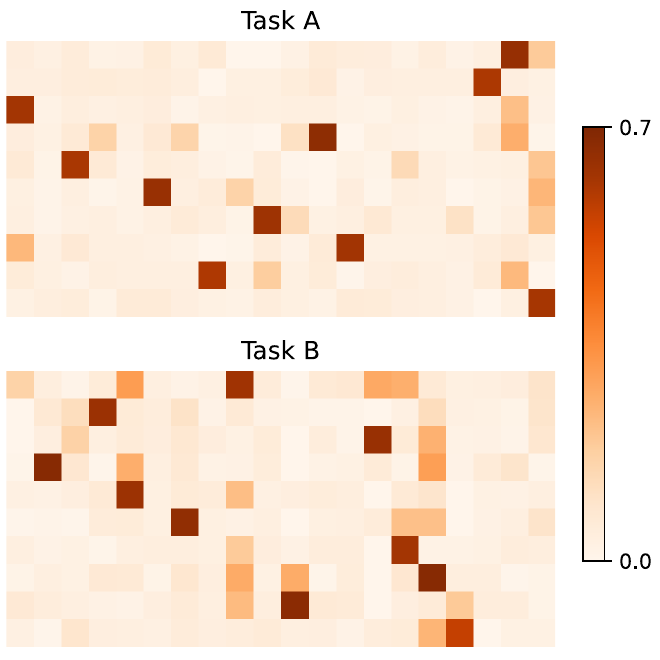}
\includeinkscape[width=0.4\textwidth]{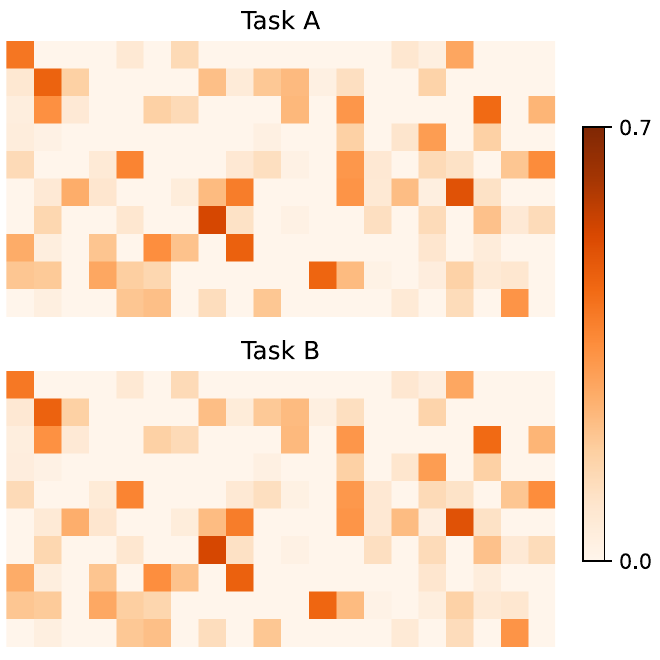}
    \caption{\textbf{Regularized, but not non-regularized, fully-connected network specializes single neurons in each row as `gates' per task and exhibits specificity based on task.} Visualization of the unsorted second hidden layer of the flexible (\emph{left}) and forgetful (\emph{right}) fully-connected network for a single seed.}
    \label{fig:gating_deep_mono_unsorted}
\end{figure}

\begin{figure}[H]
    \centering
\includeinkscape[width=0.4\textwidth]{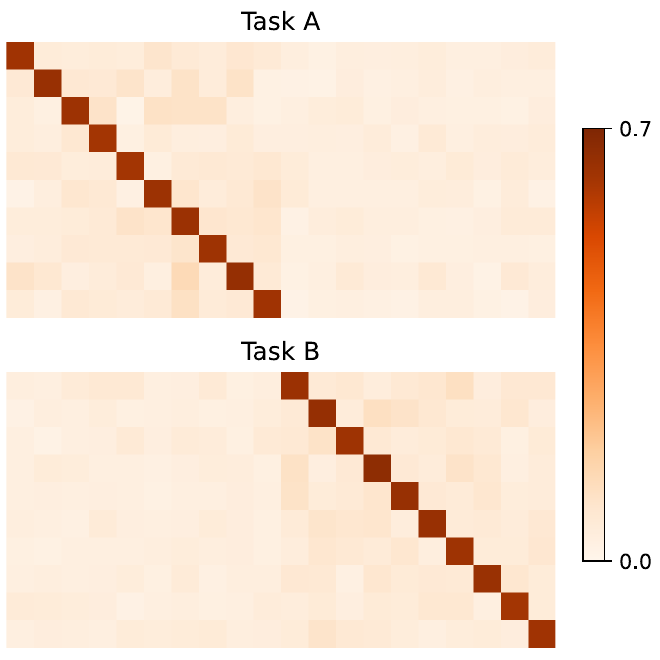}
\includeinkscape[width=0.4\textwidth]{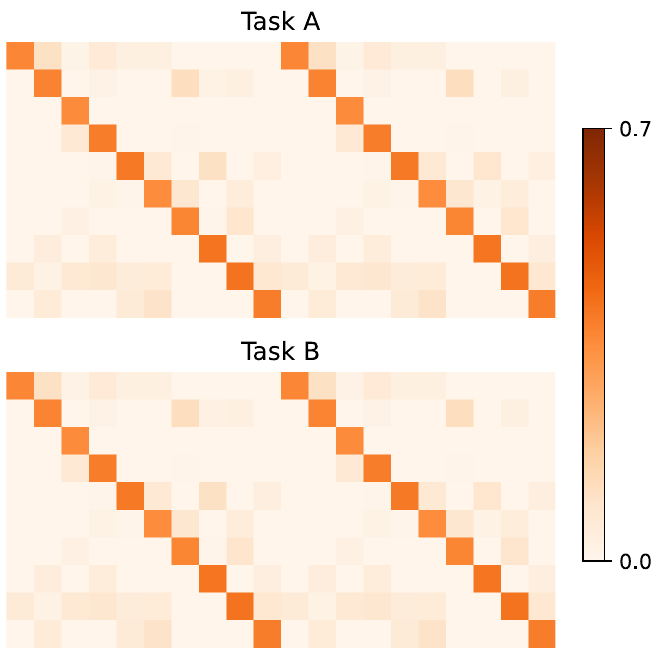}
    \caption{\textbf{Second hidden layer of regularized, but not non-regularized, fully-connected network exhibits clear task-specific gating across the diagonals of the matrix.} Visualization of the sorted second hidden layer of the flexible (\emph{left}) and forgetful (\emph{right}) fully-connected network averaged over 10 seeds.}
    \label{fig:gating_deep_mono_tenseeds}
\end{figure}

As described in the main text, we induce the flexible gating regime in a fully-connected network by applying regularization and a faster learning rate to the second layer and compare to a forgetful (unregularized) fully-connected network. Details of the sorting procedure used to identify and visualize this gating-based solution are described in \cref{app:gate_sorting}. We find that the flexible fully-connected network exhibits behavior that is qualitatively similar to the flexible NTA (\cref{fig:gating_deep_mono_metrics}). By visualizing the second layer at the end of training on different tasks in the flexible regime, we observe that the network upweights single units in each row (\cref{fig:gating_deep_mono_unsorted}A, \ref{fig:gating_deep_mono_tenseeds}A), which act as gates for the first layer rows. Instead, in the forgetful regime, the network has multiple upweighted units in each row and the units do not change behavior across different tasks, exhibiting a lack of task-specificity and gating-like behavior (\cref{fig:gating_deep_mono_unsorted}B, \ref{fig:gating_deep_mono_tenseeds}B).

\newcommand{\myone}{\scriptscriptstyle{(1)}}
\newcommand{\mytwo}{\scriptscriptstyle{(2)}}

\subsubsection{Model specialization as a function of block size, gate timescale, and regularization strength in fully-connected network}\label{app:hyperparam}

\begin{figure}[H]
    \centering
    \includeinkscape[width=0.7\textwidth]{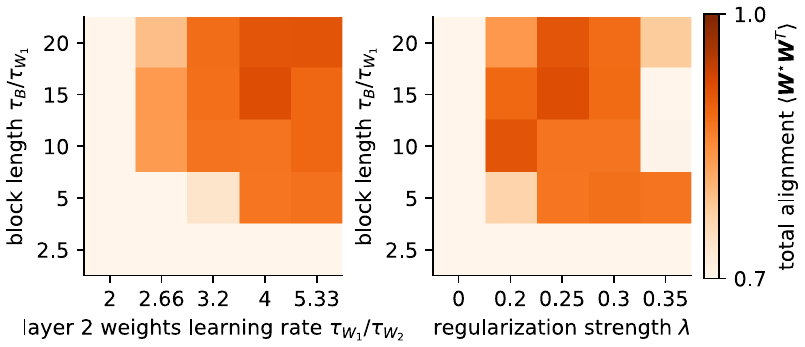}
        \caption{\textbf{Model specialization emerges as a function of block length, second hidden layer learning rate, and regularization strength in fully-connected network.} The colorbar indicates total alignment (cosine similarity) between all sets of students and teachers considered collectively.}
    \label{fig:mono_hypersearch}
\end{figure}

We perform two hyperparameter searches to illustrate the joint effects of block length, second layer learning rate, and regularization strength on the fully-connected network, similar to that we perform on the NTA model in the main text. We run the fully-connected network on each set of hyperparameters and report the total alignment of sorted teachers and students at the end of training as an overall measure of specialization, fixing all other hyperparameters (see \cref{app:hypersearch_details} for more details). We observe that the same components of block length, fast second layer learning rate, and regularization are important for specialization to emerge in the fully-connected network (\cref{fig:mono_hypersearch}), just as in the NTA model.

\subsubsection{Possibility of emergence of gating in two-layer network}
\label{app:theory_fullyconnected_gating}

In this work, we have analyzed a linear architecture with an explicit
architectural gating structure, 
\begin{equation}
y_{i}=\sum_{p=1}^{P}\sum_{j=1}^{d_{\text{in}}}c^{p}W_{ij}^{p}x_{j}=(\bm{c}\odot\bm{W}\bm{x})_i,\label{eq:lcs_hadamard}
\end{equation}

where we have notationally stacked the students $\bm{W}\coloneqq(\bm{W}^{p})_{p=1\ldots P}$
into a vector, such that $\bm{W}\in\mathbb{R}^{P\times d_{\text{out}}\times d_{\text{in}}},\bm{c}\in\mathbb{R}^{P}$.
$\odot$ here denotes the Hadamard (element-wise) product.

Prior work has considered deep linear networks \citep{saxe_mathematical_2019,Atanasov22OnsetVarianceLimited,shi_learning_2022,braun_exact_2022}, which led us to study such fully-connected network in the main text,
\begin{equation}
y_{i}=\sum_{j=1}^{d_{\text{in}}}\sum_{h=1}^{d_{\text{hid}}}W_{ih}^{\mytwo}W_{hj}^{\myone}x_{j}=(\bm{W}^{\mytwo}\bm{W}^{\myone}\bm{x})_i.\label{eq:deep_mono}
\end{equation}

The gated network considers gating as a multiplicative effect
on each output unit $i$ (or equivalently, input unit $j$), whereas
the deep network invokes an additional all-to-all weighted summation.
As such, \cref{eq:deep_mono} does not incorporate any modular structure,
yet formally resembles \cref{eq:lcs_hadamard}. To further analyze
how these settings connect, we decompose the tasks as $W_{ij}^{\star m}=\sum_{\alpha}U_{i\alpha}^{\star m}s_{\alpha}^{\star m}V_{\alpha j}^{\star m}$,
and write the student layer matrices as SVDs $W_{ih}^{\mytwo}=\sum_{\alpha}U_{i\alpha}^{\mytwo}s_{\alpha}^{\mytwo}V_{\alpha h}^{\mytwo\T}$,
$W_{hi}^{\myone}=\sum_{\alpha}U_{h\alpha}^{\myone}s_{\alpha}^{\myone}V_{\alpha j}^{\myone\T}$.
The overall model \cref{eq:deep_mono} then reads 
\begin{align*}
y_{i} & =\sum_{j}^{d_{\text{in}}}\sum_{\alpha}^{\text{min}(d_{\text{out}},d_{\text{hid}})}\sum_{\alpha'}^{\text{min}(d_{\text{hid}},d_{\text{in}})}U_{i\alpha}^{\mytwo}s_{\alpha}^{\mytwo}V_{\alpha h}^{\mytwo\T}U_{h\alpha'}^{\myone}s_{\alpha'}^{\myone}V_{\alpha'j}^{\myone\T}x_{j}\\
 & =(\bm{U}^{\mytwo}\bm{S}^{\mytwo}\bm{V}^{\mytwo\T}\bm{U}^{\myone}\bm{S}^{\myone}\bm{V}^{\myone\T}\bm{x})_{i}.
\end{align*}

If the minimum of weight dimensions $\text{min}(d_{\text{out}},d_{\text{hid}},d_{\text{in}})$
exceeds the number of task modes $\sum_{m}\text{rank}(\bm{W}^{\star m})$,
it is possible to choose/learn $\bm{V}^{\mytwo}$ and $\bm{U}^{\myone}$
such that the second layer singular values $s_{\alpha}^{\mytwo}$
effectively take the role of the gates $c^{p}$, whereas the first
layer encodes the student task representations. If we put aside the
question of learnability and only ask about expressivity, this argument
shows that a gating structure can emerge as subset of a
two-layer network. For the fully-connected model we have in the main text, $d_{\text{hid}} = 2 \, d_{\text{out}}$, giving the network the capacity to learn and remember solutions for both teachers.

\subsection{Adaptation speed}
\label{app:acceleration-NTK}
In this section, we derive the change in model output that is induced by the change in parameters depending on their configuration, thereby describing the model's adaptation speed. 
\paragraph{Neural tangent kernel}
Here, we briefly review the Neural Tangent Kernel (NTK, \citep{jacot_neural_2020}) which we then use to directly describe the adaptation speed in the output ${\bm{y}}(t)$.
For a vector-valued model $\bm{y}\in \mathbb{R}^{d_{\text{out}}} $ parameterized by a flattened parameter vector $\theta^k=(\text{flatten}(W^p_{ij},c^p))^k$,
the output evolves as
\[
\dt y_{i}=\sum_{k}\dv{y_{i}}{\theta^{k}}\dv{\theta^{k}}{t}=-\sum_{k}\dv{y_{i}}{\theta^{k}}\dv{\mathcal{L}}{\theta^{k}}=-\underbrace{\sum_{k,j}\dv{y_{i}}{\theta^{k}}\dv{y_{j}}{\theta^{k}}}_{\NTK_{ij}}\dv{\mathcal{L}}{y_{j}},
\]

where we used the chain rule and that the parameters update according
to gradient descent $\dv{\theta^{k}}{t}=-\dv{\mathcal{L}}{\theta^{k}}$
and have set the learning rate to $1$ for simplicity.

This object can be understood as a matrix operating on the output space $\mathsf{NTK}=\left(\diffd \bm{y}/\diffd \theta^{\T}\right)\left(\diffd \bm{y}^{\T}/ \diffd \theta\right)\in\mathbb{R}^{d_{\text{out}}\times d_{\text{out}}}$,
where the inner product represents the sum across parameters $\sum_{k}$
in the expression above. For the reduced model \cref{eq:dc_toy}
with $\theta=\text{flatten}( c^{p},{w}^{p}_i) _{p=1\ldots P}$, we
readily get
\[
\dv{y_{i}}{c^{p}}=w_{i}^{p},\:\dv{y_{i}}{w_{j}^{p}}=\delta_{ij}\,c^{p},
\]
where $\delta_{ij}$ is the Kronecker delta.

We then arrive at
\begin{equation}
    \NTK=\sum_{p}c^{p}c^{p}+\bm{w}^{p}\bm{w}^{p\T}, \label{eq:NTK-acc}
\end{equation}

where we adopt standard matrix notation to imply $c^{p}c^{p}\equiv c^{p}c^{p}\,\bm{I}_{d_{\text{out}}}$
as being proportional to the identity matrix $\bm{I}_{d_{\text{out}}}\in\mathbb{R}^{d_{\text{out}}\times d_{\text{out}}}$.

\paragraph{Accelerated adaptation through specialized weights and selective gates}

\begin{figure}[H]
    \centering
\includeinkscape[width=\textwidth]{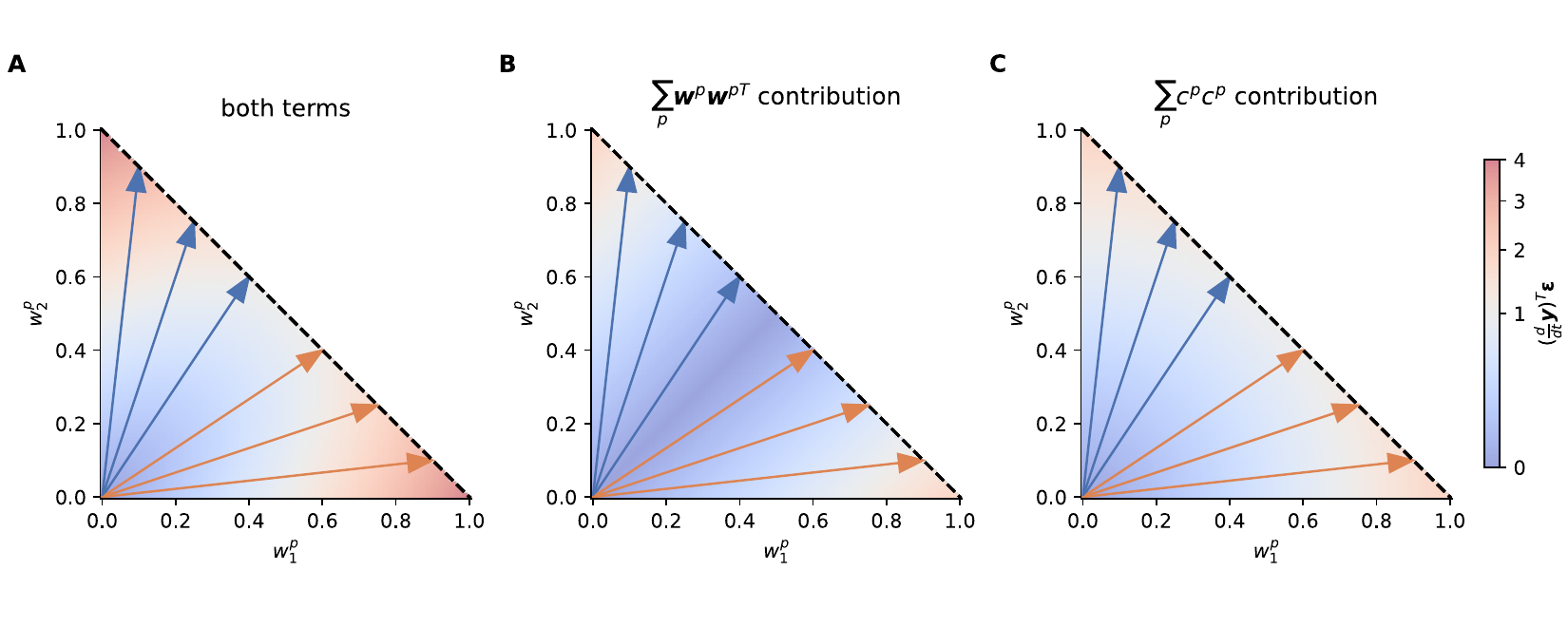}
    \caption{\textbf{Specialized students and gates accelerate adaptation}. Heatmaps of the dot product $(\frac{d}{dt} \bm{y})^\T \bm{\varepsilon}$ contributions for different terms of the Neural Tangent Kernel (NTK), depending on specialization of weight vectors $\bm{w}^1$ (\textit{blue}), $\bm{w}^2$ (\textit{orange}), of which three pairs corresponding to different degrees of specialization are shown here (pairs are formed by vectors that are symmetric along the diagonal). $c^1,c^2$ are scaled so that the sum lies on the \textit{dashed black line} (given by $L^1$ regularization). \textbf{A.} shows the total contribution of both terms of \cref{eq:NTK-acc} combined, \textbf{B.} isolates the contribution from the $\bm{w}^p \bm{w}^{p\T}$ term, and \textbf{C.} displays the contribution from the $c^p c^p$ term.  Dashed lines indicate possible solutions.}
    \label{fig:acc-NTK}
\end{figure}

To study the accelerated adaptation of the loss $\mathcal{L}_{\text{task}}=\nicefrac{1}{2}\left(\bm{y}^{\star m}-\bm{{y}}\right)^{2}$, we use the Neural Tangent Kernel of the architecture that directly describes the dynamics of the model
output. To this end, we study how the model output $\bm{y}$ changes
in response to a block switch $\bm{y}^{\star m}=(1,0)^\T\rightarrow(0,1)^\T$,
entailing an error $\be=\left(-1,1\right)^\T$ that drives a change in
model output. To see how this change reduces the loss, we calculate
its alignment with the error term as

\begin{align*}
\dt \mathcal{L}_{\text{task}} = \be^{\T}\left(\dt\bm{y}\right) & =\be^{\T}\,\mathsf{NTK}\,\be\\
 & =\be^{\T}\,\sum_{p}\left[c^{p}c^{p}+\bw^{p}\bw^{p\T}\right]\be\\
 & =\sum_{p}\left\Vert \bm{\varepsilon}\right\Vert ^{2}\left(c^{p}\right)^{2}+\left(\bw^{p\T}\be\right)^{2}.
\end{align*}

The NTK reveals that the change in output $\bm{y}$ is accelerated
through two contributions which we illustrate in \cref{fig:acc-NTK}:
First, we observed that student-teacher alignment which enters $|\bw^{p\T}\be|$ increases towards the flexible regime. We note that this acceleration however does not require unique student-teacher alignment (such that no two students match the same teacher); it is the joint effect of asymmetric gates which further facilitates unique specialization. Second, selective gates accelerate adaptation because a sparse vector with the same norm tends to have a larger sum-of-squares $\sum_p (c^p)^2$ that enters the NTK. These factors coincide in the flexible regime.

\subsection{Larger blocks enable faster specialization}\label{app:block-length}

\begin{figure}[H]
    \centering
    \includegraphics[width=\textwidth]  {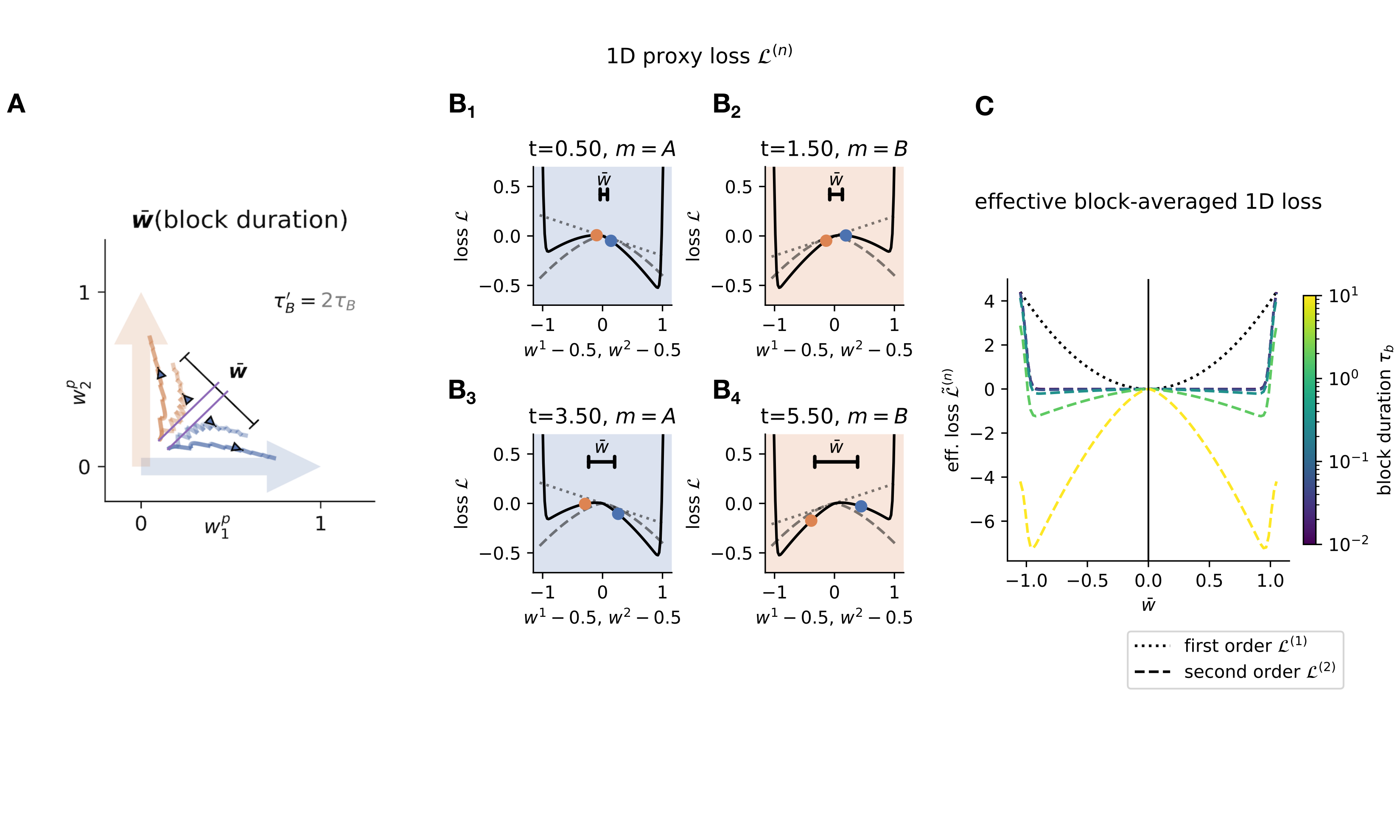}
    \caption{\textbf{Larger blocks enable faster specialization}. \textbf{A.} Two trajectories of differing block length (\textit{faint}: $\tau_B$, \textit{opaque}: $\tau_B'=2\tau_B$) of two students (\textit{blue}, \textit{orange}) in teacher subspace as in \cref{fig:mechanism}. \textbf{B.} Student dynamics in an approximate loss landscape in early learning. Subpanels 1-4 are time points in a simulation. \textit{Background}: active context $m$. The linear first-order loss does not lead to separation, as a block switch will exactly reverse any changes to specialization. In contrast, the curvature from the second order term enables students to accumulate an initial advantage in specialization. \textbf{C.} Like \textbf{B.}, but loss is in terms of the specialization variable $\bar w$. The effective loss over blocks depends on block size $\tau_B$: the longer the $\tau_B$ is, the longer the students will have to fall down the landscape in \textbf{B.}. The first-order term, corresponding to infinitely short blocks, does not prefer specialization.}
    \label{fig:block-length}
\end{figure}

Here, we calculate how the block length affects the specialization $\bar{\bm{w}}\coloneqq \bm{w}^1 - \bm{w}^2$ in students, given equal total learning time $t$. To do so, we consider periods early in learning where the students have not specialized yet, $\bar{\bm{w}}\simeq 0$, and the gates consequentially are indifferent, $\bar{\bm{c}}=0$. We then consider a period of the task, i.e. back-and-forth block switches $a\rightarrow b$, $b\rightarrow a$  that last a total of $T=2\cdot\tau_{B}$. We analyze the limit of small block sizes and ask how $\bar{\bm{w}}$ changes over $T$: for short blocks, the model is time-reversal symmetric, i.e. any change $(\dv{}{t}\bar w)\,\tau_{B}$ during $a\rightarrow b$ is exactly undone during $b\rightarrow a$. We therefore calculate the second-order effects to $\bar{\bm{w}}$ to analyze the dependence on $\tau_{B}$, where we only make an assumption on the approximate directions of $\bm{w}$ and $\be$ that apply early in learning, but leave their scale general.

\paragraph{Second-order derivatives for weight and gates}
To prepare, we first calculate the second order derivatives of the updates which we will need in what follows. By application of the product rule, we obtain for
the weights $\bw^{p}$

\begin{align*}
\dt\bm{w} & ^{p}=c^{p}\bm{\be},\\
\ddt\bm{w}^{p} & =\left(\dt c^{p}\right)\be+c^{p}\left(\dt\be\right)\\
 & =\bw^{p\T}\be\,\be-c^p \,\NTK\,\be\\
 & =\bw^{p\T}\be\,\be-c^{p}({\textstyle \sum_{p'}}c^{p'}c^{p'}+\bw^{p'}\bw^{p'\T})\be
\end{align*}

where we use $\dt\be=\dt\left(\bm{y}^{\star m}-\bm{y}\right)=-\dt\bm{y}=\dt\NTK\,\be$ within a block.

For the gates $c^{p}$ , we get

\begin{align*}
\dt c^{p} & =\bw^{p\T}\be,\\
\ddt c^{p} & =c^p\be^{\T}\be-\bw^{p \T}\,\NTK\,\be.
\end{align*}

We then take the difference of the weight derivatives $\bar{\bw }= \bm{w}^1 - \bm{w}^2, \bar c = c^1 - c^2$ to get by linearity

\begin{align*}
\dt\bar{\bm{w}} & =\bar c\bm{\be}\,\rightarrow\,0\\
\ddt\bar{\bw} & =\bar{\bw}^{\T}\be\,\be - \bar c\,({\textstyle \sum_{p'}}c^{p'}c^{p'}+\bw^{p'}\bw^{p'\T})\be\,\rightarrow\,\bar{\bw}^{\T}\be\,\be
\end{align*}

where we assume that the gates have approximately equal values $\bar c\approx0$
when no specialization has taken place yet.

\paragraph{Effect on specialization}

The sum of second derivatives after having seen a switch $a\rightarrow b$
and $b\rightarrow a$ is subsequently:

\begin{align*}
\ddt\bar{\bw}\Bigr|_{a\rightarrow b}+\ddt\bar{\bw}\Bigr|_{b\rightarrow a} & =\bar{\bm{w}}^{\T}\left(\be^{b}\,\be^{b}+\be^{a}\,\be^{a}\right)\\
 & =\left\Vert \bar{\bm{w}}\right\Vert \left\Vert \be\right\Vert \left(\be^{b}-\be^{a}\right)\\
 & =2\left\Vert \bar{\bm{w}}\right\Vert \left\Vert \be\right\Vert \,\be,
\end{align*}

where we use that $\bar{\bm{w}}$ is the component parallel with the
error signal, implying $\bar{\bw}^{\T}\be^{p}=\left\Vert \bar{\bm{w}}\right\Vert \left\Vert \be^{p}\right\Vert $
for the errors $\be^{b},\be^{a}=\pm\nicefrac{1}{2}\,C\,(1,\,-1)^{\T}$
for some constant $C$ depending on weight magnitude, and small $\bar{\bm{w}}(0)$. We herein assumed that the errors between blocks only differ by a sign, $\be\coloneqq\be^{b}=-\be^{a}$. In particular, we neglect the change in $c^p$ for simplicity. Note that this is a good approximation only if the block size $\tau_B$ is much shorter than the timescales of $\tau_c$. While this limits quantitative predictions of this approximation for the setting we consider, we expect that it identifies the qualitative mechanism.

Introducing the period $T$, which is double the block length $\tau_{B}=T/2$ (spanning two blocks of length $\tau_B$),
we get 
\begin{align}
\bar{\bw}(T=2\cdot\tau_{B})= & \bar{\bw}(0)+2\cdot\frac{1}{2}\left(2\left\Vert \bar{\bm{w}}\right\Vert \left\Vert \be\right\Vert \,\be\right)\tau_{B}^{2}
\label{eq:wbar-within-block}
\end{align}
where the factor $\frac{1}{2}$ in the first line is due to being
at second order in the Taylor expansion of the update. Setting $\bar{\bm{w}}(0)=0$,
this means that the cumulative change of two periods together lasting
$t=2T$ (i.e. spanning two pairs of blocks totaling four blocks) is 
\begin{align}
\bar{\bm{w}}(t\:=\:\text{\ensuremath{2} periods of \ensuremath{T\:=\:2\times\left(2\cdot\tau_{B}\right)}}))=2\times2\cdot\frac{1}{2}\left(2\left\Vert \bar{\bm{w}}\right\Vert \left\Vert \be\right\Vert \,\be\right)\tau_{B}^{2}=2\left(2\left\Vert \bar{\bm{w}}\right\Vert \left\Vert \be\right\Vert \,\be\right)\tau_{B}^{2}
\label{eq:wbar-across-block}
\end{align}

In contrast, doubling the block size $\tau_{B}'=2\tau_{B}$ thus ($T'=2T$),
but running for the same amount of time $t$ (i.e., two blocks of double the block length $\tau_B$) gives 
\[
\bar{\bm{w}}(t\:=\:\text{\ensuremath{1} period of \ensuremath{T'\:=\:1\times\left(2\cdot2\tau_{B}\right)}})=1\times2\cdot\frac{1}{2}\left(2\left\Vert \bar{\bm{w}}\right\Vert \left\Vert \be\right\Vert \,\be\right)\left(2\tau_{B}\right)^{2}=4\left(2\left\Vert \bar{\bm{w}}\right\Vert \left\Vert \be\right\Vert \,\be\right)\tau_{B}^{2},
\]

which is twice as large as the short-block version. This explains
why larger blocks can lead to faster specialization. 

\paragraph{Mechanical analogy}
The importance of the quadratic term for specialization can be understood through a mechanical analogy for the weights as particles: gradient flow corresponds to the dynamics of particles undergoing an overdamped Newtonian motion in a potential. To this end, we consider the proxy loss potential $\mathcal{L}^{(n)}$ that induces the respective first- and second-order time dynamics described by \cref{eq:wbar-within-block} when considering gradient flow $\tau_w \dt w = -\nabla_w \mathcal{L}^{(n)}(w)$. The resulting loss potential $\mathcal{L}^{(n)}$ is polynomial and grows $\propto -w$ (first-order) and $\propto -|w|^{3/2}$ (second-order), as can be verified by plugging in a solutions $w(t)\propto t$, $w(t)\propto t^2$. 

To first-order, block changes will result in exactly opposite gradients, which will revert any changes from the previous block due to the lack of momentum effects in gradient flow (\cref{fig:block-length}B, \textit{dotted line}). In contrast, the quadratic term in the time dynamics can be understood as resulting from an effective non-linear loss potential, breaking this time-reversal symmetry between blocks (\cref{fig:block-length}B, \textit{dashed line}). Note that the sum $\mathcal{L}^{(1)} + \mathcal{L}^{(2)}$ may still be monotonic.

When aggregating this effect over many block changes, it gives rise to an \textit{effective} loss potential $\tilde{\mathcal{L}}$ for the specialization variable $\bar{w}$ (\cref{fig:block-length}C). As the first-order terms cancel out over blocks, the effective loss potential does not have a preferred specialized configuration. When including second-order terms however, the preferred state becomes specialized. Moreover, the speed of this specialization depends on the timescale of the blocks $\tau_B$: the larger $\tau_B$, the further the particles increase their advantage down the non-linear loss potential in \cref{fig:block-length}B, which manifests as a steeper and thereby faster double-well loss potential in \cref{fig:block-length}C.

\subsection{Exact solutions under the condition of symmetry}\label{app:exact-solutions}

\subsubsection{Solving the differential equations under the condition of symmetric specialization}

The multiplicatively-coupled dynamics allow for emergent specialization of the students for one of the teachers. Due to the regularization on $c$ as well as the orthogonality of the teachers, these dynamics are competitive, meaning that the students will, after a number of task switches, each specialize for a different teacher. We saw before that the most rapid adaptation occurs when the system is in this state where each teacher is matched by exactly one student. We formalize this state more generally under a condition of \textit{symmetric specialization}, in which $\bar w = \bar w_1 = w_1^1 - w_1^2 = w_2^2 - w^1_2 = \bar w_2$. In this section, we present exact solutions to the learning dynamics that occur in this system under this condition. We assume without loss of generality that path $p$ in the NTA student is aligning with teacher $m$. We will see that these solutions closely match the adaptation dynamics of the full NTA model.

\begin{figure}[htbp]
    \centering
    \includegraphics[width=\textwidth]{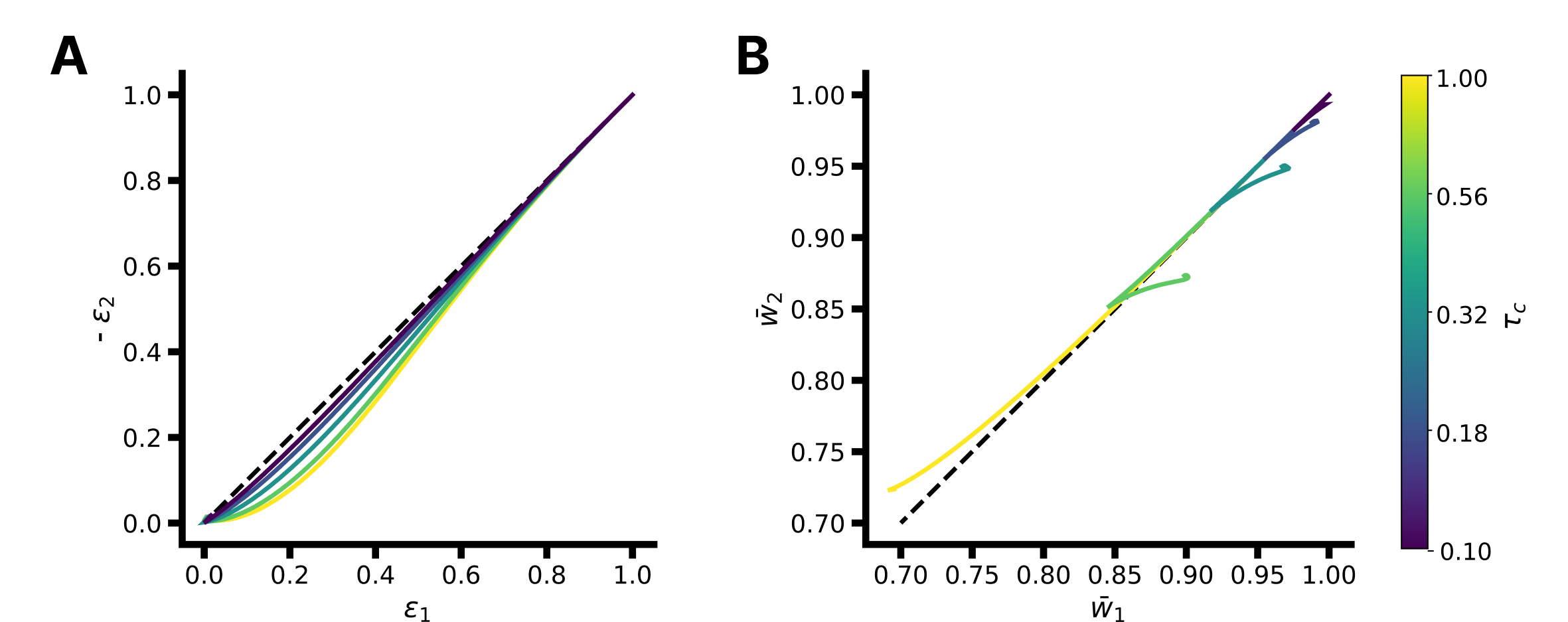}
    \caption{\textbf{Accuracy of assumptions used in calculating the exact solutions, only one of which is required.} \textbf{A.} Accuracy of the assumption $\bar \varepsilon = \varepsilon_1 = - \varepsilon_2$. \textbf{B.} Accuracy of the assumption $\bar w = w^1_1 - w^2_1 = w^2_2 - w^1_2$.}
    \label{fig:assumptions}
\end{figure}

The learning updates of the two specialization components are $\dv{\bar w_1}{t} = \bar c \, \varepsilon_1$ and $\dv{\bar w_2}{t} = - \bar c \, \varepsilon_2$. Therefore, the symmetric specialization condition is inherently preserved when  $\varepsilon_1 = - \varepsilon_2$. We calculate the accuracy of both relationships in \cref{fig:assumptions}.

\subsubsection{Learning occurs dominantly within but also outside of the specialization subspace over the course of a single block}

\label{sec:solvingexactsolssupp}

The differential equation \cref{eqn:shietaldiffeq} then results from dividing the two update equations for $\dv{c}$ and $\dv{\bm{w}}$. We can then solve this differential equation to obtain the relationship

\begin{equation}
\tau_c \, \bar{c}^2 =  2\, \tau_w \, \bar{w}^2 + C
\end{equation}
\noindent where $C$ is an integration constant, which we determine by plugging in the initial conditions $\bar c = \bar w = 1$ to represent the theoretical ideal for the flexible regime.

\subsubsection{Learning in the orthogonal component}

\label{sec:cocomponent}

As is highlighted by the fact that the analytical solution is not traversed in full in~\cref{fig:mechanism}E, learning also occurs outside of the specialization subspace. This learning can be characterized by co-specialization which is characteristic of the forgetful regime

\begin{equation}
\bar{\bar{w}} \coloneqq \frac{(w_1^1 - w_2^1) + (w^2_1 - w_2^2)}{2}.
\label{eq:defdbarw}
\end{equation}

\begin{figure}[H]
    \centering
    \includegraphics[width=\textwidth]{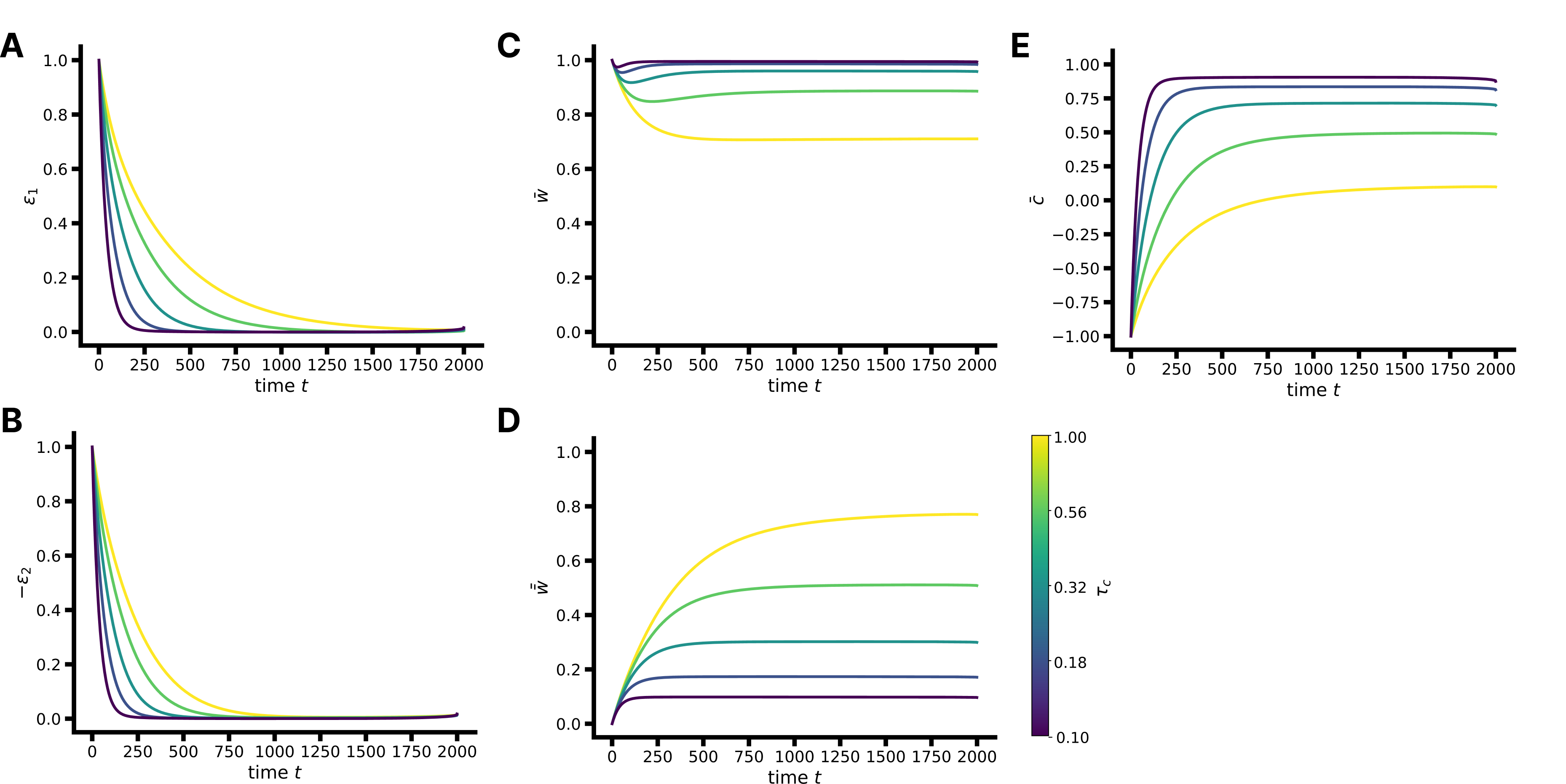}
    \caption{\textbf{Learning occurs within and outside of the specialization subspace.} \textbf{A.} First component of the error following a task switch from task A to task B for different values of $\tau_c$. \textbf{B.} Second component of the error across the same timeframe. \textbf{C.} Adaptation of weight matrices in the specialization space. \textbf{D.} Orthogonal component of learning that measures adaptation of both teachers for the current task. \textbf{E.} Gate change in the specialization subspace.}
    \label{fig:appdx_specialization_block}
\end{figure}

\cref{fig:appdx_specialization_block} shows learning along both components $\bar w$ and $\bar{ \bar w}$, the two error components $\varepsilon_1$ and $\varepsilon_2$, as well as the separation of the gates $\bar c$ over the course of a single block beginning from the same initial conditions used in the differential equation.

\subsection{Gated model generalizes to perform task and subtask composition}\label{app:generalization}

\begin{figure}[H]
    \centering
    \includegraphics[width=\textwidth]{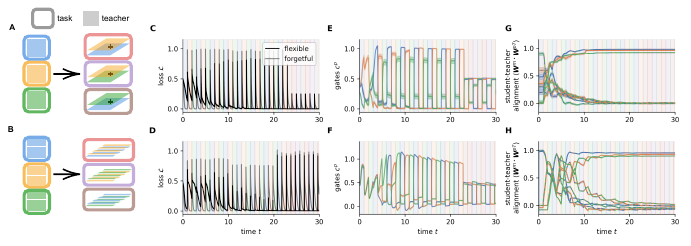}
    \caption{\textbf{Gated model generalizes to compositional tasks.} \textbf{A.} Task composition consists of new tasks that sum sets of teachers previously encountered. \textbf{B.} Subtask composition consists of new tasks that concatenate alternating rows of sets of teachers previously encountered. Loss (\textbf{C.},\textbf{D.}), gating activity (\textbf{E},\textbf{F.}), and student-teacher alignment (\textbf{G.},\textbf{H.}) of models on generalization to task composition (\emph{top}) and subtask composition (\emph{bottom}).}
    \label{fig:appdx_generalization}
\end{figure}

As stated in the main text, we consider two settings to evaluate whether the gate layer can recombine previous knowledge for compositional generalization. We first train the NTA model with three paths on three teachers A, B, and C individually, and then change the network on \textbf{task composition} (\cref{fig:appdx_generalization}A) or \textbf{subtask composition} (\cref{fig:appdx_generalization}B). Task composition proceeds the same as our standard setup.

In subtask composition, to allow our model the possibility to compose not only tasks (i.e., gate the entire student matrix $\bm{W}^p$), but also subtasks (individual rows of $\bm{W}^p$), we increase the expressiveness of our gating layer by using an independent gate for each neuron (or row of $\bm{W}^p$) in the student hidden layer and allow gradient descent to update these gates individually. We call this the per-neuron gating version of our model. 

In principle, the per-neuron gating NTA has $Pd_{\text{out}}$ independent paths modulated by gates. Thus, in order to study whether specialization and gating occurs for each teacher, we sort the $Pd_{\text{out}}$ paths into $P$ paths of size $d_{\text{out}}$. We do this by computing the cosine similarity between each row in the first layer $\bm{W}$ and the teachers $\bm{W}^{\star}$. We then sort the rows of the first layer to align with the rows of the rows of the teachers that they best match, such that we identify their respective students to visualize student-teacher alignment (\cref{fig:appdx_generalization}H). Additionally, we permute the gating layer $\bm{c}$ to match this sorting. We take the mean of the sorted gates for each student to visualize the task-specific gating (\cref{fig:appdx_generalization}F). 

We find that both the per-student and per-neuron gating NTA models can solve task and subtask generalization tasks, respectively, and maintain their learned specialization after transitioning to compositional settings (\cref{fig:appdx_generalization}C-H). We additionally observe that the gating variables learn to appropriately match the latent structure of the generalization tasks by turning off the non-contributing gate and evenly weighting the two ``on'' gates. We again observe the rapid adaptation of the flexible NTA to compositional tasks compared to the forgetful regime. 

\subsection{Non-orthogonal teachers}\label{app:nonortho_teachers}
\begin{figure}[H]
    \centering
    \includegraphics[width=\textwidth]{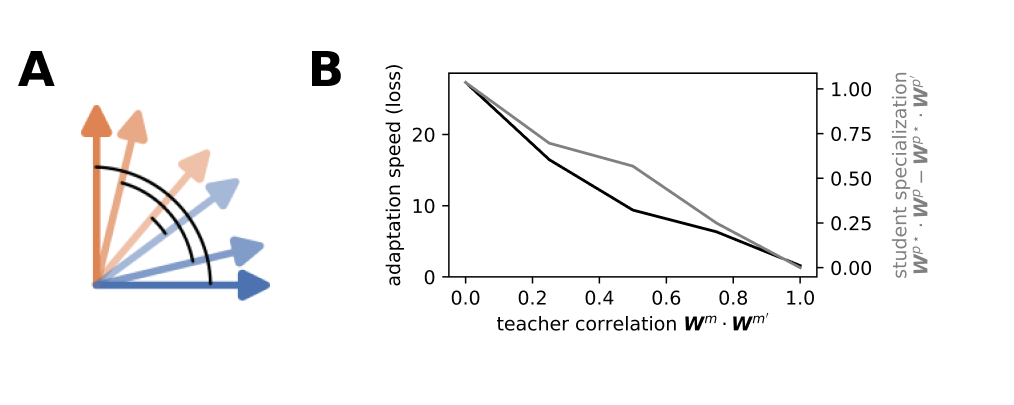}
    \caption{\textbf{Robustness to relaxing orthogonality between teachers.} \textbf{A.} Illustration of changing teacher cosine similarity. \textbf{B.} Adaptation speed as measured by the loss after a block switch (\textit{black}) and student specialization (\textit{gray}), both as a function of the teacher similarity. 0 represents the orthogonal case studied in the main text.}
    \label{fig:rebuttal_teacher_ortho}
\end{figure}

\begin{figure}[H]
    \centering
    \includegraphics[width=\textwidth]{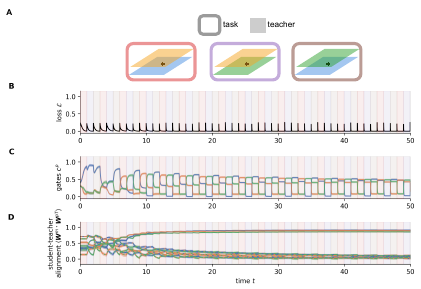}
    \caption{\textbf{Flexible NTA successfully specializes to underlying teachers even when trained on non-orthogonal tasks.} \textbf{A.} Tasks are created by adding different pairs of teachers, such that each task is non-orthogonal to every other task. \textbf{B.} Loss during learning. \textbf{C.} Gating variables learn to appropriately match the latent structure of the tasks. \textbf{D.} Students learn to specialize to teacher components, despite the non-orthogonality of the tasks.}
    \label{fig:nonorthog_tasks}
\end{figure}

In the main text, we have worked with the assumption that different tasks are approximately orthogonal to permit our theoretical analysis. This assumption holds for randomly-generated teachers when the input dimension is high. In simulations, we implemented this condition by constructing the corresponding rows of teachers to be orthogonal, $\bm{w}_i^{\star 1} \cdot \bm{w}_i^{\star 2} = 0$. Still, it is unclear what happens when the teachers are not even approximately orthogonal. We investigate this question empirically in \cref{fig:rebuttal_teacher_ortho} and find that specialization decays gracefully as the orthogonality assumption is relaxed.

We also design a set of three non-orthogonal tasks using three orthogonal teachers, where each task is created by adding different pairs of the teachers (\cref{fig:nonorthog_tasks}A). Thus, every task has some similarity (and is non-orthogonal) with every other task. We find that the flexible NTA can successfully solve these tasks and identify their underlying latent teacher structure, learning to specialize and gate all teacher components which comprise the overall set of tasks (\cref{fig:nonorthog_tasks}B-D).

\subsubsection{Experiments on fashionMNIST dataset}

\begin{figure}[H]
    \centering
    \includegraphics[width=\textwidth]{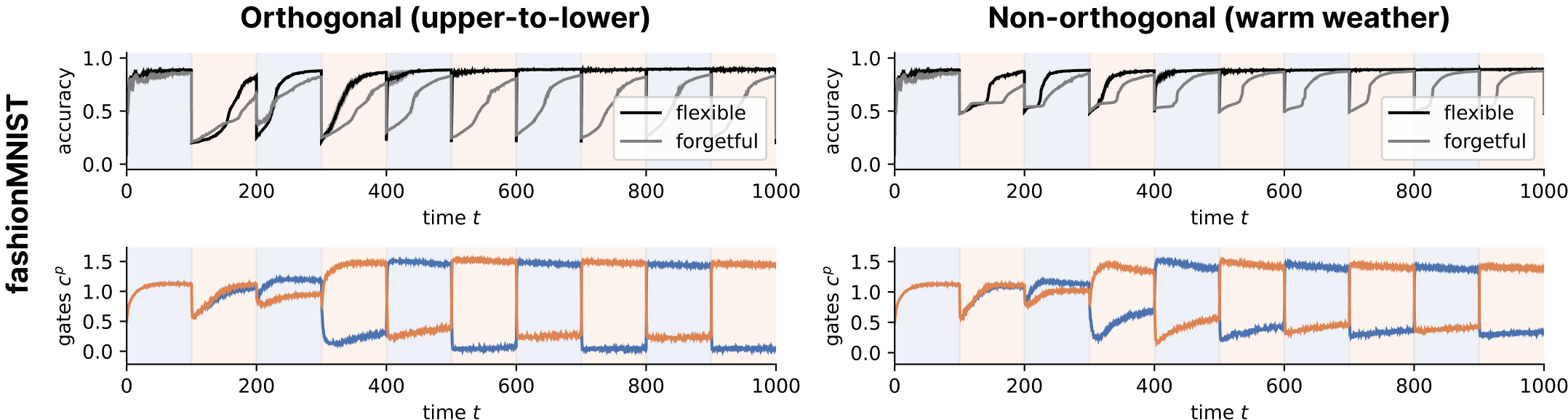}
    \caption{NTA quickly adapts across fashionMNIST for (\textit{left}) an orthogonal sorting based on upper-to-lower items of clothing and (\textit{right}) a correlated sorting for warm-to-cold weather clothing. The panels show (\textit{top}) accuracy on the test set and (\textit{bottom}) activity of the gates. We show mean and standard error with 10 seeds.}
    \label{sfig:fashion_mnist}
\end{figure}

We explicitly compare the network's performance on two different versions of fashionMNIST based on tasks that might appear in a real-world setting. The original fashionMNIST dataset has items sorted roughly by order of commonality, with the label 0 being assigned to T-shirts, and the label 9 being assigned to ankle boots. We generate two different permutations of these labels representing other real-world sorting of the items that have different amounts of shared structure with the original. The close-to-orthogonal ordering sorts the clothing from upper to lower body, and orders the labels 0, 2, 4, 6, 8, 1, 3, 5, 7, 9. The ordering with more shared structure represents warm to cold weather clothing, and orders the labels to 0, 1, 5, 3, 7, 6, 2, 4, 8, 9.
The results show that the stereotypical NTA-like task switching behavior and specialization emerges for both settings at a similar speed despite baseline performance being higher on the task with shared structure (\cref{sfig:fashion_mnist}).

\subsection{Few-shot adaptation in the low sample rate regime}\label{app:rebuttal_samples}
\begin{figure}[H]
    \centering
    \includegraphics[width=\textwidth]{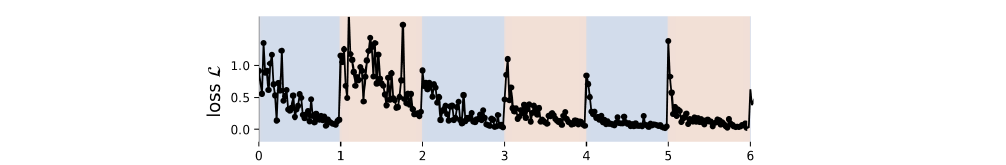}
    \caption{\textbf{Few-shot adaptation after block switches.} Like \cref{fig:main}A in the main text, but with coarsely discretized time to examine the adaptation after a single sample. As this drastically reduces signal-to-noise ratio, we average over 100 samples. Markers indicate a single step of gradient descent on one sample.}
    \label{fig:rebuttal_samples}
\end{figure}
In the main text, we have considered the case of large batch size (or equivalently, small time discretization) that allows taking a sample average when going towards the theoretical, equivalent model. This averaging reduces noise in the gradient signal stemming from random samples, so that it is unclear whether learning is still possible when sample rate is low. As theoretical analysis is challenging for this case, we investigate this empirically in \cref{fig:rebuttal_samples} and find that the qualitative phenomenon is preserved even if only a single sample $B=1$ is used for every gradient update.

\section{Technical details}\label{app:technical-detail}
\subsection{Notation}\label{app:notation}

\begin{table}[H]
\caption{Overview of notation used throughout the paper.}
\centering
\resizebox{1.\textwidth}{!}{  %
\begin{tabular}{@{}ll@{}}
\toprule
\textbf{Symbol}                & \textbf{Description}             \\ \midrule
$\bm{x}^b \in \mathbb{R}^{d_{\text{in}}},\ b = 1...B$ & input sample of a batch \\ \\
$\bm{y}^b \in \mathbb{R}^{d_{\text{out}}},\ b = 1...B$ & model output \\ \\
$\bm{y}^{\star m} \in \mathbb{R}^{d_{\text{out}}}, \ m = 1...M \text{ or } a, b, ...$  & target label in task $m$ \\ \\
$\bm{\varepsilon}=\bm{y}^{\star} - \bm{y}\in \mathbb{R}^{d_{\text{out}}}$ & prediction error \\ \\

$c^p \in \mathbb{R},\ p = 1...P$ & gates for each pathway $p$ \\ \\
$\bm{W}^p \in \mathbb{R}^{d_{\text{out}} \times d_{\text{in}}}$ & student weights for each pathway $p$ \\ \\
$\bm{W}^{\star m} \in \mathbb{R}^{d_{\text{out}} \times d_{\text{in}}},\ m = 1...M \text{ or } a, b, ...$ & teacher weights for each task $m$ \\ \\
$\bm{w}^p \equiv \bm{w}^p_\alpha \in \mathbb{R}^2$ & 2D vector for reduced model  (for each teacher singular value $\alpha$) \\ \\
$W_{ij} = \sum_\alpha^{\text{min}(d_{\text{out}}, d_{\text{in}})} U_{i\alpha} s_\alpha V_{\alpha j}^\T$ & singular value decomposition of weights \\ \\
$\tau_w=\eta_w^{-1}$, $\tau_c=\eta_c^{-1}$ & parameter time scale (inverse learning rate) \\ \\
$\tau_B$ & block length \\ \\
$\mathcal{L} = \mathcal{L}_{\text{task}} + \mathcal{L}_{\text{reg}}$ & loss \\ \\
$\bar{w}_1 = w_{m=1}^{p=1} - w_{m=1}^{p=2}$ & specialization for teacher 1\\ \\
$\bar w_2 = w_{m=2}^{p=2} - w_{m=2}^{p=1}$ & specialization for teacher 2 \\ \\
$\bar{w} = \frac{1}{2} (\bar w_1 + \bar w_2)$ & overall specialization \\ \\
$\bar{c} = c^1 - c^2$ & separation of gates \\ \\ 
$\bar{\bar{w}} = ({(w_{m=1}^{p=1} - w_{m=2}^{p=1}) + (w^{p=2}_{m=1} - w_{m=2}^{p=2})})/{2}$ & unspecialized learning \\ \\
\\\bottomrule
\end{tabular}
}
\end{table}

\subsection{Hyperparameters}\label{app:hyperparam_list}
We perform the gradient calculations and the simulation of gradient flow using the JAX framework and make our implementation publicly available at \href{https://github.com/aproca/neural_task_abstraction}{https://github.com/aproca/neural\_task\_abstraction}. 

Our hyperparameters for all experimental settings are listed in the table below. We use $\mathcal{L}_{\text{norm-L1}}$ for most experiments, except for generalization to subtask composition and the flexible fully-connected network experiments where we use $\mathcal{L}_{\text{norm-L2}}$ (see \cref{app:regularization} for a discussion on this choice). For the cases where we induce the forgetful regime as an experimental control, we use the same hyperparameters, set regularization to 0 ($\lambda_{\text{nonneg}}, \lambda_{\text{norm-L1}}, \lambda_{\text{norm-L2}}=0$), and may or may not adjust the learning rate of the gating layer. Differences in hyperarameters from the main model and control are denoted in the tables below as `main / control.'

\begin{table}[H]
\caption{Hyperparameters.}
\label{apptable:hyperparams}
\centering
\resizebox{1.\textwidth}{!}{  %
\begin{tabular}{@{}lllllll@{}}
\toprule
Hyperparameter                 & \makecell{Task specialization \\ (\cref{fig:main})} & \makecell{Task composition \\ (\cref{fig:generalization},\ref{fig:appdx_generalization})} & \makecell{Subtask composition \\ (\cref{fig:generalization},\ref{fig:appdx_generalization})} & \makecell{Reduced model \\ (\cref{fig:mechanism})} & \makecell{Fully-connected network \\(\cref{fig:gating_deep_mono_sorted_gates},\ref{fig:gating_deep_mono_metrics},\ref{fig:gating_deep_mono_unsorted},\ref{fig:gating_deep_mono_tenseeds}) }               & \makecell{MNIST \\(\cref{fig:mnist})}               \\ \midrule
$P$            & 2     & 3  & 3 & 2 & 2 & 2 \\
$M$            & 2     & 3  & 3 & 2 & 2 & 2 \\
$d_{\text{in}}$            & 20     & 20  & 20 & 1 & 20 & 64\\
$d_{\text{hid}}$            &     &   &  & & 20 & \\
$d_{\text{out}}$            & 10     & 6  & 6 & 2 & 10 & 10\\
$\lambda_{\text{nonneg}}$ & 0.091 / 0    & 0.5 / 0 &  0.023 / 0 & 0.091 / 0 & 0.2 / 0 & 0.5 / 0 \\
$\lambda_{\text{norm-L1}}$ & 0.456 / 0    & 1.25 / 0  &  0 & 0.455 / 0 & 0 & 0.25 / 0\\
 $\lambda_{\text{norm-L2}}$& 0& 0& 0.011 / 0& 0 & 0.1 / 0&0\\
$\tau_w$ & 1.3     & 0.2  & 0.2 & 5 & 0.06 & 10\\
$\tau_c$ & 0.03 / 1.3    & 0.03  & 0.005 & 0.7 & 0.01 & 0.005 / 10\\
batch size           & 200     & 200  & 200 & 200 & 200 & 100\\
seeds           & 10     & 10  & 10 & 1 & 10 & 10 \\
number of blocks $n$ & 20 & 30 & 30  & 17 & 30 & 10 \\
$\tau_B$       & 1     & 1  & 1 & 1 & 1 & 1\\ %
$dt$ & 0.001     & 0.001  & 0.01 & 0.001 & 0.01 & 0.001\\\bottomrule
\end{tabular}
}
\end{table}

\begin{table}[H]
\caption{Hyperparameters II.}
\label{apptable:hyperparams2}
\centering
\resizebox{1.\textwidth}{!}{  %
\begin{tabular}{@{}llllll@{}}
\toprule
Hyperparameter                 & NTA hyperparameter search (\cref{fig:hypersearch}) & Fully-connected hyperparameter search (\cref{fig:mono_hypersearch}) & Task switching (\cref{fig:task_switching_speed}) & Non-orthogonal tasks (\cref{fig:nonorthog_tasks}) & Non-orthogonal teachers (\cref{fig:rebuttal_teacher_ortho})               \\ \midrule
$P$            & 2     & 2  & 2 & 3 & 2 \\
$M$            & 2     & 2  & 2 & 3  & 2\\
$d_{\text{in}}$            & 20     & 20  & 20 & 20 & 20\\
$d_{\text{hid}}$            &      & 20  &  &  & \\
$d_{\text{out}}$            & 10     & 10  & 10 & 6 & 10\\
$\lambda_{\text{nonneg}}$ & 0.5     & 0.23  & 0.18 / 0& 0.33 & 0\\
$\lambda_{\text{norm-L1}}$ & 1.25     & 0  & 0.36 / 0 & 0.83 & 0\\
 $\lambda_{\text{norm-L2}}$& 0& 0.11 & 0 & 0 & 0.5\\
$\tau_w$ & 0.1     & 0.04 & 0.07 & 0.05 & 0.016 \\
$\tau_c$ & 0.005     & 0.01 & 0.01 & 0.03 & 0.016 \\
batch size           & 200     & 200 & 200 & 200 & 200 \\
seeds           & 10     & 10 & 10 & 10  & 1\\
number of blocks $n$ & 7 & 20 & 30 & 50  & 10\\
$\tau_B$       & 1     & 1 & 1 & 1 & 1\\ %
$dt$ & 0.001     & 0.01 & 0.01 & 0.001 & 0.01 \\\bottomrule
\end{tabular}
}
\end{table}

\begin{table}[H]
\caption{Hyperparameters III.}
\label{apptable:hyperparams3}
\centering
\resizebox{1.\textwidth}{!}{  %
\begin{tabular}{@{}llllll@{}}
\toprule
Hyperparameter                 & Full vs. reduced model (\cref{fig:full-vs-toy}) & Slow high-d students (\cref{fig:dimensionality})    & Redundant paths (\cref{fig:repr_cost})    & Few-shot adaptation (\cref{fig:rebuttal_samples})   & fashionMNIST (\cref{sfig:fashion_mnist})      \\ \midrule
$P$           & 2  & 2  & 4 & 2 & 2 \\
$M$           & 2  & 2  & 2 & 2 & 2 \\
$d_{\text{in}}$            & 20 / 1 & 30 & 20 & 20 & 64\\
$d_{\text{out}}$           & 10 / 2 & 30 & 10 & 10 & 10 \\
$\lambda_{\text{nonneg}}$      & 0.091  & 0.091 & 0.194 / 0.545 & 0.091 & 0.5 / 0\\
$\lambda_{\text{norm-L1}}$      & 0.455  & 0.455 & 0.968 / 2.727  & 0 & 0 \\
 $\lambda_{\text{norm-L2}}$ & 0 & 0 & 0 & 0.455 & 0.25 / 0 \\
$\tau_w$     & 1.3  & 0.5 & 1.3 & 1 & 10 \\
$\tau_c$     & 0.03 / 0.06 & 0.1 & 0.03 & 0.01 & 0.005 / 10 \\
batch size       & 200  & 200 & 200 & 1 & 100 \\
seeds       & 1  & 10 & 1 & 10 & 10 \\
number of blocks $n$ & 20 & 20 & 20 & 6 & 10 \\
$\tau_B$        & 1  & 1 & 1 & 1 & 1 \\ %
$dt$     & 0.001  & 0.001 & 0.001 & 0.02 & 0.001 \\\bottomrule
\end{tabular}
}
\end{table}

\subsection{Regularization}\label{app:regularization}

\begin{figure}[htbp]
    \centering
    \includegraphics[width=0.7\textwidth]{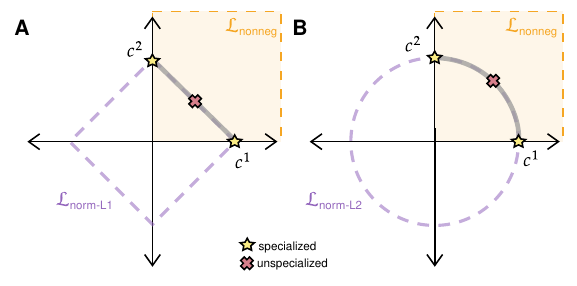}
    \caption{\textbf{The effect of regularization on gating variables.} Regularization encourages competition between gates while preventing degeneracy of solutions. Importantly, regularization does not force gating variables to be specialized, as illustrated by the red $\times$. This holds for two regularizers we consider, \textbf{A.} $\mathcal{L}_{\text{norm-L1}}$ and \textbf{B.} $\mathcal{L}_{\text{norm-L2}}$.}
    \label{fig:regularization_visual}
\end{figure}

We use a combined regularizer that is motivated by biological constraints
on our model parameters. The regularizers alleviate the underspecification of
the solution space of our linear model 
and facilitate symmetry breaking to allow the model to specialize different components, while not forcing specialization (\cref{fig:regularization_visual}).
Here, we detail the effect of these regularizing
terms. To this end, recall the definition of the reduced model, \cref{eq:reduced_model}

\[
\bm{y}=c^{1}\bm{w}^{1}+c^{2}\bm{w}^{2}.
\]

\paragraph*{Nonnegative neural activity}
The gating variables steer the model output multiplicatively. Biologically, such an interaction is mediated by a firing rate, an inherently positive variable. Computationally, this has implications on the solution space: with random initialization, almost all configurations of $\bm{w}^{1}$
and $\bm{w}^{2}$ form a basis of $\mathbb{R}^{2}$. By definition,
this means that there will always be two coefficients $c^{1}$, $c^{2}$$\in\mathbb{R}$
that will yield the correct solution. Nonnegativity constrains this
set to lie in the positive quadrant of a 2D space. In particular,
\begin{align*}
    & \mathcal{L}_{\text{nonneg}} = \sum_{p=1}^P \max (0, -c^p)
\end{align*}

\paragraph*{Alleviating invariance via competition}

Even in the desired specialized configuration, the model
is invariant under 
\[
c^{p},\bm{W}^{p}\rightarrow ac^{p},\bm{W}^{p}/a
\]

for any scalar $a$. We hence bound the norm of the vector $\bm{c}=\left(c^{1},c^{2}\right)^{\T}$.

We consider two regularizers based on the $L^1$ and $L^2$ norm
\begin{align*}
    & \mathcal{L}_{\text{norm-L1}} = \nicefrac{1}{2}(\|\bm{c}\|_1 - 1)^2 \\
    & \mathcal{L}_{\text{norm-L2}} = \nicefrac{1}{2}(\|\bm{c}\|_2 - 1)^2
\end{align*}
$\mathcal{L}_{\text{norm-L1}}$ encourages sparsity in the gates which is beneficial when there are few active gating variables as in the NTA model (with a single teacher active in each task). However, if we consider cases with potentially many active gating variables, as in the per-neuron gating NTA (see \cref{app:generalization}) or a deep fully-connected network (or even many teachers active at once), favoring sparsity restricts expressivity of the model. In these cases, we instead use $\mathcal{L}_{\text{norm-L2}}$. In practice, both regularizers facilitate specialization robustly across many settings.

Applying nonnegativity and norm regularization together has the effect of inducing competition between gating variables. While there is a solution where gates are equal in magnitude (as shown in \cref{fig:regularization_visual}), deviating from this solution while minimizing regularization loss will cause one gate to increase in magnitude and the other to decrease. Thus, this competitive effect facilitates symmetry-breaking in the gates.

Finally, we note that although these regularizers facilitate symmetry breaking through competition, they simultaneously allow for compositionality such that multiple gates can be active at once. We show this in several experiments, namely our studies of nonorthogonal tasks (\cref{app:nonortho_teachers}), compositional generalization (\cref{app:generalization}), and fully-connected networks (\cref{app:deep-monolithic}).

\subsection{Description of metrics used across experiments}\label{app:metrics}

\paragraph{Student-teacher alignment}
We compute a metric of alignment of each student and each teacher to determine whether students are specializing and, if so, to which teachers they specialize. We do this by computing the similarity between each student $\bm{W}^p$ and teacher $\bm{W}^{\star m}$. More specifically, we take the mean of the cosine similarity between student and teacher row vectors. We then sort each student and its gate $c^p$ to the teacher it has the highest cosine similarity with.

\paragraph{Total alignment}
We compute a metric of total alignment of the network students and teachers to evaluate overall specialization. After computing student-teacher alignment and sorting each student to its respective teacher, we concatenate all students and teachers and compute the overall cosine similarity between the set of students and teachers.

\subsubsection{Description of sorting performed in per-neuron NTA and deep fully-connected network} \label{app:gate_sorting}

In the cases of more expressive models, such as the per-neuron NTA and fully-connected network, there are $Pd_{\text{out}}$ and $d_{\text{out}} \times Pd_{\text{out}}$ respective independent paths modulated by gates. Thus, in order to study whether specialization and gating occurs for each teacher, we sort these into $P$ paths. To do this, we compute the cosine similarity between each row in the first layer $\bm{W}$ and each row in the teachers $\bm{W}^{\star}$. We then sort the rows of the first layer to align with the rows of the teachers that they best match. 

Additionally, we permute the second layer to match this sorting. In the per-neuron NTA, this corresponds to scalar gates that are multiplied to each row. In the fully-connected network, this corresponds to the columns of the second layer. Finally, we take the mean of the sorted gates (set of $d_{\text{out}}$ columns for the fully-connected network) for each student to visualize teacher-specific gating.

\subsection{Hyperparameter search} \label{app:hypersearch_details}
\paragraph{NTA} For the two hyperparameter searches we perform, we run the NTA model on each set of hyperparameters and report the total alignment of concatenated teachers and students at the end of training as an overall measure of specialization. We fix all other hyperparameters.

When varying gate learning rate and block length, we fix the regularization strength to $\lambda_{\text{nonneg}}=0.5$, $\lambda_{\text{norm-L1}}=1.25$. When varying regularization strength, we fix gate learning rate $\tau_w/\tau_c=20$. The regularization strength $\lambda$ is multipled separately for each type of regularizer such that $\lambda_{\text{nonneg}}=5\lambda/3$ and $\lambda_{\text{norm-L1}} = 25\lambda/6$. 

\paragraph{Fully-connected network} We also perform two hyperparameter searches on the fully-connected network. We run the fully-connected network on each set of hyperparameters and report the total alignment of sorted teachers and students at the end of training as an overall measure of specialization. We fix all other hyperparameters.

When varying second layer learning rate and block length, we fix the regularization strength to $\lambda_{\text{nonneg}}=0.23$, $\lambda_{\text{norm-L2}}=0.11$. When varying regularization strength, we fix gate learning rate $\tau_{W^{\mytwo}}/\tau_{W^{\myone}}=4$. The regularization strength $\lambda$ is multiplied separately for each type of regularizer such that $\lambda_{\text{nonneg}}=10\lambda/11$ and $\lambda_{\text{norm-L2}} = 5\lambda/11$.

\subsection{Flexible fully-connected network}
We randomly generate two orthogonal teachers $\boldsymbol{W}^{\star m} \in \mathbb{R}^{d_{\text{out}} \times d_{\text{in}}}$. We initialize our fully-connected networks to have two weight layers, $\boldsymbol{W}^{\myone} \in \mathbb{R}^{2d_{\text{out}} \times d_{\text{in}}}$ and $\boldsymbol{W}^{\mytwo} \in \mathbb{R}^{d_{\text{out}} \times 2d_{\text{out}}}$. We use a faster learning rate and regularize the second layer during training. We treat each weight ${W}^{\mytwo}_{ij}$ as a gate that enters into the regularization terms described in \cref{app:regularization}. Student-teacher alignment, total alignment, and gate sorting is then performed as described above.

\subsection{MNIST}
\label{app:mnist}

The original convolutional network features a single convolutional layer with three feature maps and kernel size four, followed by a MaxPool layer of kernel size 2, a ReLU nonlinearity, a and fully-connected sigmoid, ReLU and log softmax layers of size 512, 64, and 10 respectively. The networks are trained using cross-entropy loss with a one-hot encoding for labels. The NTA portion is then trained beginning from the hidden layer representations of the final hidden layer with 64 units.
hyperparameters for the NTA portion are given in \cref{apptable:hyperparams} for MNIST and \cref{apptable:hyperparams3} for fashionMNIST.